\documentclass[11pt]{article}

\usepackage[preprint]{acl}

\usepackage{times}
\usepackage{latexsym}

\usepackage[T1]{fontenc}
\usepackage[utf8]{inputenc}

\usepackage{microtype}

\usepackage{inconsolata}

\usepackage{graphicx}
\usepackage{microtype}
\usepackage{subcaption}
\usepackage{booktabs} % for professional tables
\usepackage{enumitem}
\usepackage{hyperref}
\usepackage{amsmath}
\usepackage{amssymb}
\usepackage{mathtools}
\usepackage{amsthm}
\usepackage{caption}
\usepackage{algorithm}
\usepackage{algorithmic}
\usepackage{multirow}
\usepackage{makecell}

\usepackage{tcolorbox}
\tcbuselibrary{breakable, skins}

\definecolor{promptbg}{RGB}{245, 247, 250}
\definecolor{promptborder}{RGB}{100, 130, 170}
\title{IntroConformal: Conformal Factuality Guarantees for Large Vision-Language Models via Introspective Signals}

\author{Md. Atabuzzaman \qquad Christian Alexander 
 \qquad Chris Thomas \\
Department of Computer Science \\ Virginia Tech \\
\texttt{\{atabuzzaman, cmalexander, christhomas\}@vt.edu}}

\begin{document}
\maketitle

\begin{abstract}
    Large Vision--Language Models (LVLMs) have achieved strong multimodal performance, yet ensuring the factual correctness of generated content remains challenging. Existing methods that provide statistical guarantees on factuality typically rely on external verifiers or generation-time confidence signals, which introduce auxiliary dependencies or often fail for confident but incorrect outputs. We argue that reliable factuality control can instead be achieved through \textit{introspective} signals derived from the model itself. We introduce \textbf{IntroConformal}, a training-free Conformal Risk Control (CRC) framework that provides finite-sample, distribution-free factuality guarantees. We first instantiate it with layer-wise semantic stability, a conformity score derived from hidden-state representations, and then propose verification probability, a stronger score capturing the model's self-administered judgment on claim factuality. Across multiple LVLM architectures, IntroConformal satisfies the conformal risk guarantee while substantially reducing abstention and achieving competitive or superior claim-level discrimination relative to external verifier–based baselines.\footnote{Code and Dataset: \url{https://github.com/Atabuzzaman/Introconformal}}
\end{abstract}

%%%%%%%%%%%%%%%%%%%%%%%%%%%%%%%%%%%%%%%%%%%%%%%%%%%%%%%%%%%%

\section{Introduction}
\label{sec:intro}

LVLMs have achieved remarkable progress across vision--language tasks and are increasingly deployed in high-stakes domains such as medical reporting and autonomous systems. However, these models remain prone to generating content that is not factually grounded in the input image. Such failures are particularly concerning because users often have no reliable way to distinguish incorrect outputs from correct ones: confident yet non-factual generations can appear highly plausible, undermining trust and limiting real-world deployment~\citep{zhang2024vl, li2025towards}.

Addressing this challenge requires more than heuristic mitigation; it calls for formal, finite-sample bounds on the rate of non-factual claims while preserving useful model outputs. Although many approaches attempt to mitigate factual errors through prompting strategies, decoding heuristics, or auxiliary verification, most do not offer formal statistical guarantees. Recently, conformal prediction and CRC have emerged as promising tools for uncertainty quantification in large language models (LLMs), providing finite-sample, distribution-free guarantees on error rates~\citep{vovk2005algorithmic, angelopoulosuncertainty, bates2021distribution, quachconformal2024, cherian2024large}. When applied to LVLMs, these methods can bound factuality risk at user-specified levels, for example targeting a 10\% error rate by filtering non-factual claims under factuality control protocols~\citep{li2025towards}.

However, existing conformal factuality frameworks for LVLMs suffer from a fundamental \emph{signal bottleneck}. They define conformity scores using either generation-time token log-probabilities or external verification models~\citep{quachconformal2024, li2025towards}. Generation-time confidence is often unreliable, as models can remain highly confident even when factually ungrounded~\citep{xiongcan, cheninside}, while external verifiers introduce additional dependencies and complicate deployment in resource-constrained settings.

In contrast, we argue that reliable factuality control can be achieved through introspective signals derived from the model itself, without external verifiers or auxiliary supervision. Prior work shows that non-factual generation is associated with internal inconsistencies, including layer-wise semantic drift and unstable hidden-state trajectories~\citep{azaria2023internal, cheninside, zhang2025icr, nie2025mechanistic, bu2026sampling}. Although these signals arise during a single forward pass, they are largely ignored by conformal approaches that treat LVLMs as black boxes~\citep{li2025towards}.

Building on this observation, we introduce \textbf{IntroConformal}, a training-free Conformal Risk Control (CRC) framework that provides finite-sample, distribution-free factuality guarantees using conformity scores derived entirely from the model itself. We first instantiate it with \emph{layer-wise semantic stability} ($S_{\text{sem}}$), which measures alignment between mid- and late-layer hidden-state representations on claim tokens. While $S_{\text{sem}}$ satisfies the CRC guarantee across architectures, its discrimination between factual and non-factual claims remains modest (Table~\ref{tab:signal_quality_mscoco}), resulting in high abstention rates that limit practical utility. To address both limitations, we propose \emph{verification probability} ($S_{\text{prob}}$), a stronger conformity score that queries the same model with a binary factuality prompt and reads the output logits rather than sampling a discrete answer. Across multiple LVLM architectures, $S_{\text{prob}}$ reduces abstention and improves claim-level discrimination over both $S_{\text{sem}}$ and external verifier--based baselines (Table~\ref{tab:mscoco_main_results}), while preserving the conformal risk guarantee. Our main contributions are:

\begin{itemize}[noitemsep, topsep=0pt, leftmargin=*]
    \item We propose \textbf{IntroConformal}, a training-free CRC framework for LVLM factuality that derives conformity scores from the model itself, without external verifiers or auxiliary supervision.
    \item We introduce two conformity scores: \textbf{layer-wise semantic stability}, capturing cross-layer hidden-state alignment, and \textbf{verification probability}, which queries the same model with a binary factuality prompt and reads the output logits.
    \item Across multiple LVLM architectures and benchmarks, IntroConformal satisfies the CRC guarantee, reducing abstention and improving F1 over external verifier-based baselines, and achieving higher claim-filtering efficiency and response accuracy than decoding-based methods.
\end{itemize}

%%%%%%%%%%%%%%%%%%%%%%%%%%%%%%%%%%%%%%%%%%%%%%%%%%%%%%%%%

\section{Related Work}
\label{sec:related_work}

\textbf{Uncertainty and Hallucination in LLMs.}
Uncertainty estimation and hallucination detection in LLMs have been extensively studied. Early approaches based on verbalized confidence and sampling-based consistency~\citep{kuhnsemantic, xiongcan} require multiple generations and fail on confident hallucinations~\citep{cheninside}, while semantic entropy methods~\citep{kuhnsemantic, nikitin2024kernel, duan2024shifting} require repeated sampling at inference time. More recent work shows that internal activations encode factuality signals through hidden-state classifiers and representation geometry~\citep{han2024semantic, li2025semantic}. Most closely related to ours, mechanistic interpretability studies reveal that non-factual generations manifest as layer-wise semantic drift and unstable hidden-state trajectories~\citep{azaria2023internal, chuang2024dola, cheninside, zhang2025icr, bu2026sampling}; however, these approaches remain primarily diagnostic and lack distribution-free statistical guarantees.

%%%%-------------------------------------

\noindent
\textbf{Uncertainty in LVLMs.}
Uncertainty estimation in LVLMs introduces additional multimodal grounding challenges. Several approaches focus on selective prediction under insufficient visual context~\citep{liu2024detecting, lau2025uncertainty, khan2024consistency} or address inconsistency through cycle-consistency and attention-alignment frameworks~\citep{shah2019cycle, selvaraju2020squinting}, while perturbation-based methods have shown mixed results compared to representation-based signals~\citep{avestimehr2025detecting}. These approaches typically rely on heuristics or auxiliary models and lack formal statistical guarantees---a gap our work addresses through conformal risk control with introspective signals.

%%%%-----------------------------------------

\noindent
\textbf{Conformal Prediction for Factuality Control.}
Conformal prediction provides distribution-free, finite-sample guarantees for uncertainty quantification~\citep{vovk2005algorithmic, angelopoulosuncertainty, bates2021distribution}. Recent applications to LLMs include multiple-choice tasks~\citep{ye2024benchmarking}, open-ended generation~\citep{quachconformal2024}, and claim-level filtering for LVLM factuality using learned~\citep{vishwakarmaprune} or external scoring functions~\citep{li2025towards}. However, these methods define conformity using generation-time token probabilities or external verifiers such as CLIP~\citep{clip}, which are unreliable for confident non-factual generations or require auxiliary models. Our work bridges conformal prediction with mechanistic interpretability~\citep{cheninside, zhang2025icr, bu2026sampling} by defining training-free conformity scores derived entirely from the model itself, yielding finite-sample, distribution-free factuality guarantees without external verifiers or generated token probabilities.

%%%%%%%%%%%%%%%%%%%%%%%%%%%%%%%%%%%%%%%%%%%%%%%%%%%%%%%%%%%%%

\begin{figure*}[t]
    \centering
    \includegraphics[width=\linewidth]{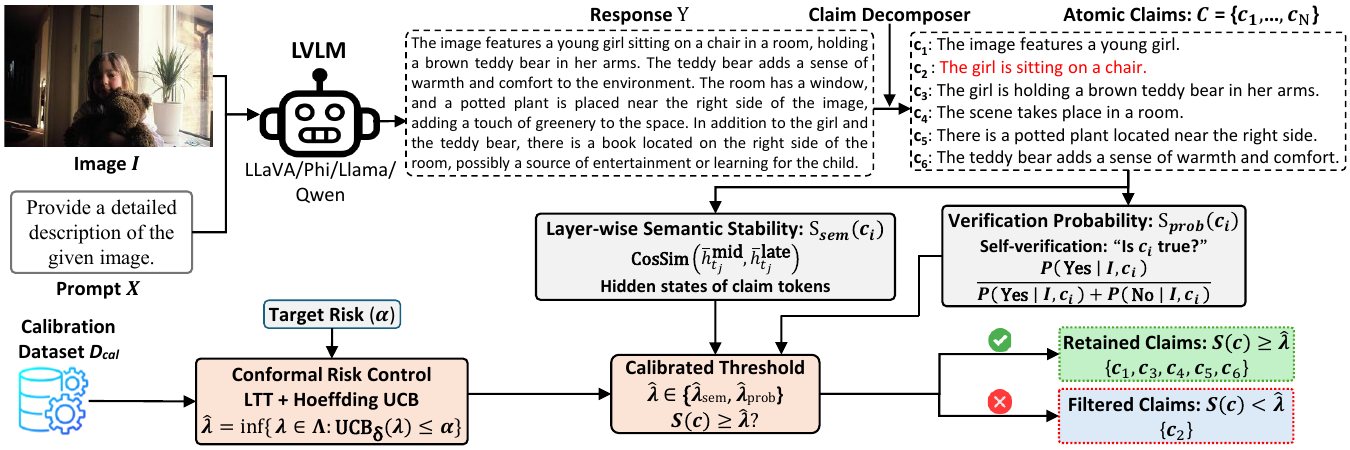}
    \vspace{-0.75cm}
    \caption{Overview of \textbf{IntroConformal}. Our framework decomposes an LVLM response into atomic claims (non-factual claim $c_2$ highlighted in \textcolor{red}{red}) and computes two conformity scores: layer-wise semantic stability $S_{\text{sem}}$ from hidden-state representations and verification probability $S_{\mathrm{prob}}$ from the model's \texttt{Yes}/\texttt{No} factuality judgment; the user selects one score for calibration and filtering. Conformal risk control then calibrates a threshold $\hat{\lambda} \in \{\hat{\lambda}_{\mathrm{sem}}, \hat{\lambda}_{\mathrm{prob}}\}$ on $\mathcal{D}_{\mathrm{cal}}$, retaining claims with $S(c) \geq \hat{\lambda}$ and providing finite-sample, distribution-free factuality guarantees.}
    \vspace{-0.5cm}
    \label{fig:overview}
\end{figure*}

%%%%%%%%%%%%%%%%%%%%%%%%%%%%%%%%%%%%%%%%%%%%%%%%%%%%%%%%%%%%%

\section{Method}
\label{sec:method}

We introduce IntroConformal, a framework for statistically controlling non-factual generation risk in LVLMs via \emph{introspective conformity scores} (Figure~\ref{fig:overview}), where \textit{introspective} refers to scores derived from the same LVLM without external verifiers or auxiliary supervision. Unlike prior conformal approaches that rely on generation-time token probabilities~\citep{quachconformal2024} or external verifiers~\citep{li2025towards}, IntroConformal derives both scores directly from the model itself. We propose two such scores: \emph{layer-wise semantic stability} ($S_{\text{sem}}$), measuring hidden-state alignment across layers, and \emph{verification probability} ($S_{\text{prob}}$), capturing the model's binary factuality judgment. We first instantiate the CRC framework with $S_{\text{sem}}$, then show that $S_{\text{prob}}$ improves discrimination and reduces abstention while preserving the guarantee.

%%%%%-------------------------------------------------

\subsection{Problem Formulation}
\label{sec:problem}
Given an image $I$, a textual prompt $X$, and a model response $Y$, we decompose $Y$ into atomic, verifiable claims $\mathcal{C} = \{c_1, \ldots, c_N\}$ following prior work~\citep{li2025towards}. Our goal is to retain a subset $\hat{\mathcal{C}} \subseteq \mathcal{C}$ whose factuality is statistically controlled while providing response-level risk guarantees. Let $\mathcal{L}(c, I) \in \{0, 1\}$ denote a non-factuality indicator, where $\mathcal{L}(c, I) = 1$ if claim $c$ is not supported by image $I$ and $0$ otherwise. Formally, we construct a selection rule such that the expected rate of non-factual claims among the retained set is bounded by a user-specified risk level $\alpha \in [0, 1]$:
\begin{equation}
\mathbb{E}\left[\frac{1}{|\hat{\mathcal{C}}|} \sum_{c \in \hat{\mathcal{C}}} \mathcal{L}(c, I)\right] \leq \alpha,
\label{eq:risk}
\end{equation}
with risk defined as zero when $|\hat{\mathcal{C}}| = 0$. This objective aligns with CRC, which provides finite-sample, distribution-free guarantees for selection-conditional risk. We next define our two scores.

%%%%-------------------------------------------------

\subsection{Layer-wise Semantic Stability}
\label{sec:ssem}
Our first conformity score captures semantic stability across the model's internal representations. Prior work shows that non-factual generation is accompanied by semantic drift in the final layers, where representations diverge from those formed at intermediate decoding stages~\citep{cheninside, wang2025mllm, bu2026sampling}. Let $h_t^{(\ell)} \in \mathbb{R}^d$ denote the hidden state of token $t$ at layer $\ell$. We define two disjoint layer sets: $\mathcal{M}$, comprising the $K_{\text{mid}} = 8$ transformer layers immediately preceding $\mathcal{T}$, and $\mathcal{T}$, comprising the final $K_{\text{late}} = 4$ layers. These values are fixed across all architectures; a sensitivity analysis is provided in Appendix~\ref{sec:effect_layer}.

For each claim token $t_j$, we compute averaged hidden representations:
\begin{equation}
\bar{h}_{t_j}^{\text{mid}} = \frac{1}{|\mathcal{M}|} \sum_{\ell \in \mathcal{M}} h_{t_j}^{(\ell)}, \quad \bar{h}_{t_j}^{\text{late}} = \frac{1}{|\mathcal{T}|} \sum_{\ell \in \mathcal{T}} h_{t_j}^{(\ell)}.
\end{equation}
We compute the cosine similarity between these representations and average across tokens to obtain the claim-level semantic stability score:
\begin{equation}
S_{\text{sem}}(c_i) = \frac{1}{|c_i|} \sum_{j=1}^{|c_i|} \text{CosSim}\!\left(\bar{h}_{t_j}^{\text{mid}}, \bar{h}_{t_j}^{\text{late}}\right).
\end{equation}
Higher $S_{\text{sem}}$ indicates stable semantic grounding, where representations remain consistent from mid to late layers. Lower values reflect semantic drift associated with non-factual claims.

%%%%%--------------------------------------------------

\subsection{Verification Probability}
\label{sec:sprob}

While $S_{\text{sem}}$ provides a hidden-state signal of semantic consistency, its discriminative power is modest. To address this limitation, we propose \emph{verification probability} ($S_{\text{prob}}$), a stronger conformity score capturing the model's binary judgment on claim factuality.

Given an image $I$ and an atomic claim $c_i$, we prompt the same LVLM to assess whether $c_i$ is supported by $I$ via: ``Based on the image, is the following statement true? Answer with Yes or No. Statement: $c_i$.'' We apply each model's standard chat template and extract the \texttt{Yes}-token probability at the first answer position from a single forward pass, normalizing against \texttt{No} to isolate relative confidence from absolute output magnitudes:
\begin{equation}
S_{\text{prob}}(c_i) = \frac{P(\texttt{Yes} \mid I, c_i)}{P(\texttt{Yes} \mid I, c_i) + P(\texttt{No} \mid I, c_i)}.
\end{equation}
Higher $S_{\text{prob}}$ indicates greater support for the claim; lower values reflect the model's disagreement with claims extracted from its earlier response.

$S_{\text{prob}}$ is related to CoVe~\citep{cove2024} but differs in a key respect: unlike CoVe, which samples discrete verification answers and conditions further generation on them, $S_{\text{prob}}$ extracts the \texttt{Yes}-token probability directly without additional decoding steps. This makes it strictly single-pass and avoids the sampling overhead of CoVe while still conditioning explicitly on claim-image consistency rather than next-token prediction.

%%%%-----------------------------------------------------

\subsection{Conformal Risk Control}
\label{sec:crc}

To control the non-factual claim risk defined in Eq.~\eqref{eq:risk}, we adopt a Conformal Risk Control (CRC) framework based on the Learn--Then--Test (LTT) paradigm~\citep{angelopoulos2021learn, bates2021distribution}. Unlike split-conformal calibration~\citep{vovk2005algorithmic,angelopoulos2021gentle}, which calibrates a quantile of nonconformity scores to control coverage probability, CRC handles real-valued losses such as the response-level non-factual rate by selecting the least conservative threshold whose Hoeffding upper confidence bound (UCB) satisfies the target risk $\alpha$. This yields a high-probability guarantee $\mathbb{P}(R(\hat{\lambda}) \leq \alpha) \geq 1 - \delta$. We assume access to a calibration set $\{(I_k, X_k, \mathcal{C}_k, \mathcal{L}_k)\}_{k=1}^{n}$ drawn i.i.d.\ from the same distribution as test inputs, where
$\mathcal{C}_k$ denotes the set of atomic claims extracted from the model output for image-prompt pair $(I_k, X_k)$, and $\mathcal{L}_k$ provides claim-level non-factuality labels.

\noindent
\textbf{Nested claim selection.}
Given a threshold $\lambda \in \mathbb{R}$, we define a claim-level filtering operator
\begin{equation}
\hat{\mathcal{C}}_\lambda(I_k, X_k) =
\{c \in \mathcal{C}_k : S(c) \geq \lambda\},
\label{eq:filter}
\end{equation}
where $S(c)$ denotes the chosen conformity score. These sets are nested in $\lambda$, with larger thresholds inducing more aggressive filtering.

\noindent
\textbf{Per-response empirical risk.}
For each calibration example $k$, let
\begin{equation}
m_k(\lambda) \triangleq
\left|\hat{\mathcal{C}}_\lambda(I_k, X_k)\right|
\end{equation}
denote the number of retained claims. We define the response-level non-factual rate among retained claims as

\begin{equation}
\resizebox{\linewidth}{!}{$
r_k(\lambda) =
\begin{cases}
\dfrac{\sum_{c \in \hat{\mathcal{C}}_\lambda(I_k, X_k)}
\mathcal{L}(c, I_k)}{m_k(\lambda)},
& \text{if } m_k(\lambda) > 0, \\[4pt]
0,
& \text{if } m_k(\lambda) = 0.
\end{cases}
$}
\end{equation}
where the numerator counts non-factual claims among those retained. Following the selective prediction convention~\citep{geifman2017selective, li2025towards}, we
assign zero loss to a response when the model abstains by filtering all claims. Note that $r_k(\lambda) \in [0, 1]$ by construction. %\textcolor{purple}{

This fractional loss is not monotone in $\lambda$: removing a factual claim can raise the ratio. We retain it because the proportion of incorrect claims, not the absolute number, is our object of interest.

We address this via a Hoeffding concentration inequality with family-wise error rate (FWER) correction, bounding the risk below a corrected level ($\alpha^\prime \leq 0.170$) with probability ($1-\delta$), equivalently guaranteeing that at least (83\%) of retained claims are factual in expectation.

\noindent
\textbf{LTT calibration via Hoeffding UCB.}
Let $\Lambda$ denote a finite set of thresholds, taken as the unique values of $\{S(c) : c \in \mathcal{C}_k, \, k = 1, \ldots, n\}$ (optionally augmented with a value below the minimum to allow retaining all claims). For each $\lambda \in \Lambda$, we compute the empirical risk
\begin{equation}
\hat{R}(\lambda) = \frac{1}{n}\sum_{k=1}^{n} r_k(\lambda).
\end{equation}
We then construct an upper confidence bound using Hoeffding's inequality~\citep{hoeffding1963probability}:
\begin{equation}
\mathrm{Pr}(\hat{R}(\lambda) - R(\lambda) \leq -x) \leq \exp(-2nx^2),
\end{equation}
where $R(\lambda) = \mathbb{E}_{(I_k, X_k) \sim \mathcal{D}}[r_k(\lambda)]$ denotes the true expected risk over test inputs drawn from the same distribution as $\mathcal{D}_{\text{cal}}$. 

We take $x=\sqrt{\frac{\log(1/\delta)}{2n}}$ to deduce a per-$\lambda$ bound, which holds with probability $\ge 1 - \delta,$
\begin{equation}
R(\lambda) \le \mathrm{UCB}_\delta(\lambda) \triangleq \hat{R}(\lambda) + \sqrt{\frac{\log(1/\delta)}{2n}}.
\end{equation}

A per-$\lambda$ bound does not guarantee that all $\lambda \in \Lambda$ concurrently satisfy the risk constraint. Trading slightly in $\alpha$, a Bonferroni correction~\citep{angelopoulos2021learn} ensures concurrent validity: the adjustment
\begin{equation}
    \alpha' \triangleq \alpha + \left[\sqrt{\frac{\log(m/\delta)}{2n}} - \sqrt{\frac{\log(1/\delta)}{2n}}\right] 
\end{equation}
is sufficient such that if $\lambda \in \Lambda'$ satisfy $\mathrm{UCB}_\delta(\lambda) \leq \alpha$ individually, then they all concurrently satisfy
\begin{equation}
\mathrm{Pr}(R(\lambda) \leq \alpha') \geq 1 - \delta.
\end{equation}

In our experiments, with $n=400$ calibration prompts and at most $50$ claims per response, the candidate set $\Lambda$ contains $|\Lambda| \leq n \cdot 50 = 20{,}000$ unique thresholds. Testing individually at $\alpha = \delta = 0.1$ then yields concurrent risk control at $\alpha' \leq 0.170.$

We select the smallest feasible threshold satisfying the risk constraint: the infimal $\lambda \in \Lambda',$ i.e.
\begin{equation}
\hat{\lambda} = \inf\{\lambda \in \Lambda : \mathrm{UCB}_\delta(\lambda) \leq \alpha\}.
\end{equation}
This choice maximizes claim retention by selecting the least conservative threshold among concurrently valid choices. Appendix~\ref{sec:algo} (Algorithm~\ref{alg:introconformal}) summarizes the full procedure. (Note that the algorithm assumes that input $\alpha$ is to be respected, i.e. regarded as the $\alpha'$ of the derivation above.)

%%%%%%%%%%%%%%%%%%%%%%%%%%%%%%%%%%%%%%%%%%%%%%%%%%%%%%%%%%%

\begin{table*}[!ht]
\centering
\small
\setlength{\tabcolsep}{6pt}
\renewcommand{\arraystretch}{1.15}
\begin{tabular}{l l l c c c c c}
\toprule
\textbf{Task} &\textbf{Model} & \textbf{Signal} &
\textbf{Mean (F)} & \textbf{Mean (NF)} & \textbf{Difference} &
\textbf{AUROC $\uparrow$} & \textbf{$p$-value} \\
\midrule
\multirow{21}{*}{\makecell{General Scene \\Understanding \\(MSCOCO)}} 
& \multirow{4}{*}{LLaVA-1.5}
  & CLIP                & 0.2108 & 0.1911 & +0.0196 & 0.631 & $7.6 \times 10^{-33}$ \\
  & & $T_{\text{prob}}$  & 0.2723 & 0.2200 & +0.0523 & 0.611 & $1.2 \times 10^{-26}$ \\
  & & $S_{\text{sem}}$  & 0.8689 & 0.8674 & +0.0015 & 0.556 & $3.1 \times 10^{-9}$ \\
   & & $S_{\text{prob}}$ & 0.8598 & 0.6584 & \textbf{+0.2014} & \textbf{0.819} & $< 10^{-100}$ \\ \cmidrule(lr){2-8}

& \multirow{4}{*}{Phi-3.5-Vision}
  & CLIP                 & 0.2065  & 0.2029  & +0.0037  &  0.523 & $3.6 \times 10^{-2}$ \\
  & & $T_{\text{prob}}$  & 0.2444 & 0.2200 & +0.0244 & 0.555 & $8.9 \times 10^{-9}$ \\
  & & $S_{\text{sem}}$  & 0.9041 & 0.9022 & +0.0019 & 0.576 & $2.2 \times 10^{-10}$ \\
  & & $S_{\text{prob}}$ & 0.8506 & 0.6109 & \textbf{+0.2397} & \textbf{0.763} & $< 10^{-80}$ \\ \cmidrule(lr){2-8}
& \multirow{4}{*}{Llama-3.2-Vision}
  & CLIP                 & 0.2103  & 0.2067  & +0.0036  & 0.519  & $4.3 \times 10^{-2}$ \\
  & & $T_{\text{prob}}$  & 0.1015 & 0.0958 & +0.0056 & 0.531 & $3.5 \times 10^{-3}$ \\
  & & $S_{\text{sem}}$  & 0.6827 & 0.6830  & $-0.0003$ & 0.488 & $3.5 \times 10^{-1}$ \\
  & & $S_{\text{prob}}$ & 0.8312 & 0.7108 & \textbf{+0.1204} & \textbf{0.716} & $< 10^{-50}$ \\ \cmidrule(lr){2-8}

& \multirow{4}{*}{Qwen2.5-VL-7B}
  & CLIP                 & 0.2060  & 0.2037  & +0.0024  & 0.512  & $1.3 \times 10^{-1}$ \\
  & & $T_{\text{prob}}$  & 0.0203 & 0.0163 & $+0.0040$ & 0.579 & $4.7 \times 10^{-7}$ \\
  & & $S_{\text{sem}}$  & 0.7933 & 0.7921  & $+0.0011$ & 0.527 & $1.6 \times 10^{-3}$ \\
  & & $S_{\text{prob}}$ & 0.9169 & 0.7414 & \textbf{+0.1755} & \textbf{0.739} & $< 10^{-50}$ \\
  \cmidrule(lr){2-8}

& \multirow{4}{*}{Qwen3-VL-8B}
  & CLIP                 & 0.2036  & 0.2010  & +0.0026  & 0.516  & $8.5 \times 10^{-2}$ \\
  & & $T_{\text{prob}}$  & 0.1067 & 0.0802 & $+0.0265$ & 0.605 & $1.4 \times 10^{-30}$ \\
  & & $S_{\text{sem}}$  & 0.8966 & 0.8932  & $+0.0035$ & 0.566 & $2.5 \times 10^{-13}$ \\
  & & $S_{\text{prob}}$ & 0.9327 & 0.7654 & \textbf{+0.1673} & \textbf{0.699} & $< 10^{-42}$ \\

\midrule
\multirow{8}{*}{\makecell{Fine-Grained \\Captioning}} 
& \multirow{4}{*}{LLaVA-1.5}
  & CLIP              & 0.2068 & 0.1831 & +0.0236 & 0.655 & $< 10^{-50}$ \\
  & & $T_{\text{prob}}$  & 0.2794 & 0.2150  & +0.0645 & 0.652 & $< 10^{-40}$ \\
  & & $S_{\text{sem}}$  & 0.8703 & 0.8695 & +0.0008 & 0.536 & $3.7 \times 10^{-4}$ \\
  & & $S_{\text{prob}}$ & 0.8604 & 0.6756  & \textbf{+0.1849} & \textbf{0.765} & $< 10^{-100}$ \\ \cmidrule(lr){2-8}
& \multirow{4}{*}{Phi-3.5-Vision}
  & CLIP             & 0.1989 & 0.1925 & +0.0064 & 0.536 & $8.1 \times 10^{-7}$ \\
  & & $T_{\text{prob}}$  & 0.2429 & 0.2098 & +0.0331 & 0.583 & $3.1 \times 10^{-20}$ \\
  & & $S_{\text{sem}}$ & 0.9033  & 0.9025 & +0.0008 & 0.530 & $6.4 \times 10^{-3}$ \\
  & & $S_{\text{prob}}$ & 0.8220 & 0.5751 & \textbf{+0.2468} & \textbf{0.770} & $< 10^{-100}$ \\
  
\midrule

\multirow{8}{*}{\makecell{Document \\Understanding}} 
& \multirow{4}{*}{LLaVA-1.5}
  & CLIP              & 0.2452 & 0.2278 & +0.0173 & 0.631 & $< 10^{-40}$ \\
  & & $T_{\text{prob}}$  & 0.2416 & 0.2598  & $-0.0182$ & 0.493 & $9.8 \times 10^{-7}$ \\
  & & $S_{\text{sem}}$  & 0.8710  & 0.8691  & +0.0018 & 0.575 & $5.0 \times 10^{-21}$ \\
   & & $S_{\text{prob}}$ & 0.8617 & 0.7680 & \textbf{+0.0937} & \textbf{0.728} & $< 10^{-100}$ \\ \cmidrule(lr){2-8}

& \multirow{4}{*}{Phi-3.5-Vision}
  & CLIP             & 0.2434 & 0.2275 & +0.0159 & 0.597 & $2.4 \times 10^{-31}$ \\
  & & $T_{\text{prob}}$  & 0.3017 & 0.2792 & +0.0225 & 0.537 & $5.8 \times 10^{-6}$ \\
  & & $S_{\text{sem}}$  & 0.8837 & 0.8818 & +0.0019 & 0.530 & $5.7 \times 10^{-6}$ \\
  & & $S_{\text{prob}}$ & 0.8767 & 0.7621 & \textbf{+0.1147} & \textbf{0.677} & $< 10^{-50}$ \\
\bottomrule
\end{tabular}
\vspace{-0.25cm}
\caption{Signal quality across all three tasks and LVLM architectures 
on the calibration set. For each signal we report the mean score on 
factual (F) and non-factual (NF) claims, their difference, AUROC, and 
Welch's $t$-test $p$-value. CLIP (CLIP-ViT-Large) is the external verifier used by 
CONFLVLM~\citep{li2025towards}; $T_{\text{prob}}$ is the average 
log-probability of claim tokens $c_i$ force-decoded under the 
verification prompt context, serving as a token-confidence baseline 
distinct from the Yes/No judgment of $S_{\text{prob}}$; $S_{\text{sem}}$ 
and $S_{\text{prob}}$ are our proposed conformity scores. $S_{\text{prob}}$ consistently achieves the strongest discrimination across all tasks and architectures.}
\label{tab:signal_quality_mscoco}
\vspace{-0.5cm}
\end{table*}

%%%%%%%%%%%%%%%%%%%%%%%%%%%%%%%%%%%%%%%%%%%%%%%%%%%%%%%%%%

\section{Experiments and Evaluation}
\label{sec:experiments}
We evaluate IntroConformal on three vision--language generation tasks requiring grounded factual generation: general scene understanding, fine-grained captioning, and document understanding. We design our experiments to assess two key questions: (i) whether signals extracted from the model itself meaningfully separate factual and non-factual claims, and (ii) whether these signals enable valid and efficient conformal risk control under finite-sample guarantees. Following CONFLVLM~\citep{li2025towards}, we evaluate both response-level conformal risk and claim-level diagnostic metrics across multiple LVLM architectures and datasets.

%%%%----------------------------------

\subsection{Experimental Setup}
\label{sec:experi_setup}
We evaluate IntroConformal on three representative vision--language benchmarks. For general scene understanding, we use the MSCOCO-based benchmark introduced by CONFLVLM~\citep{li2025towards}, consisting of 500 images (400 calibration, 100 test) with claim-level factuality annotations. For fine-grained captioning, we construct a balanced benchmark using CUB~\citep{cub}, Stanford Cars~\citep{cars}, and Stanford Dogs~\citep{dogs} by selecting one image per category, resulting in 516 images (400 calibration, 116 test). For document understanding, we use invoice images from SROIE~\citep{huang2019icdar2019}, randomly selecting 500 images following the same 400/100 calibration--test split. We evaluate five LVLM architectures: LLaVA-1.5-7B~\cite{llava}, Phi-3.5-Vision-Instruct~\cite{phi3}, Llama-3.2-11B-Vision~\cite{llama3}, Qwen2.5-VL-7B-Instruct~\cite{Qwen2.5-VL}, and Qwen3-VL-8B-Instruct~\cite{qwen3}.

Following CONFLVLM~\citep{li2025towards}, we decompose model responses into atomic claims and annotate for factual correctness with respect to the input image. Annotation reliability is established at two levels. For general scene understanding, we directly use the publicly available CONFLVLM annotations, where GPT-4o~\cite{gpt4o} labels were validated against human raters with an Intra-class Correlation Coefficient (ICC) of 0.85, indicating strong inter-rater reliability. For fine-grained captioning and document understanding, claim decomposition is performed using GPT-4o-mini and factuality labels are generated using GPT-5.4. To assess reliability, one human annotator independently reviewed 372 claims (54.3\% factual, 45.7\% non-factual by GPT label) across 50 randomly selected images, achieving 86.0\% agreement with GPT-5.4 labels and Cohen's $\kappa$ of 0.71, indicating substantial inter-annotator agreement~\citep{landis1977measurement}. Together, these results confirm strong alignment between automatic and human factuality judgments. Appendix~\ref{sec:prompts} presents claim decomposition and annotation prompts.

%%%%%%%%%%%%%%%%%%%%%%%%%%%%%%%%%%%%%%%%%%%%%%%%%%%%%%%%%%%

\begin{table*}[!ht] %t
\centering
\small
\setlength{\tabcolsep}{6pt}
\renewcommand{\arraystretch}{1.15}
\begin{tabular}{l l l c c c c c}
\toprule
\multirow{2}{*}{\textbf{Task}} &\multirow{2}{*}{\textbf{Model}} & \multirow{2}{*}{\textbf{Method}} & \multicolumn{2}{c}{\textbf{Response-level}} & \multicolumn{3}{c}{\textbf{Claim-level}} \\
\cmidrule(lr){4-5} \cmidrule(lr){6-8}
& & &
\textbf{Risk $\downarrow$} & \textbf{Abst. $\downarrow$} &
\textbf{TPR $\uparrow$} & \textbf{Precision $\uparrow$} & \textbf{F1 $\uparrow$} \\

\midrule
\multirow{21}{*}{\makecell{General Scene \\Understanding \\(MSCOCO)}} 
& \multirow{4}{*}{LLaVA-1.5}
  & CONFLVLM             & 0.102 & 57\% & 0.953 & 0.343 & 0.504 \\
  & & $T_{\text{prob}}$  & 0.045 & 74\% & \textbf{0.981} & 0.354 & 0.520 \\
  & & $S_{\text{sem}}$   & \textbf{0.030} & 64\% & \textbf{0.981} & 0.366 & 0.533 \\
  & & $S_{\text{prob}}$  & 0.054 & \textbf{25\%} & 0.974 & \textbf{0.414} & \textbf{0.581} \\ \cmidrule(lr){2-8}
& \multirow{4}{*}{Phi-3.5-Vision}
  & CONFLVLM           & 0.094 & 65\% & 0.945 & 0.254 & 0.401 \\
  & & $T_{\text{prob}}$  & 0.068 & 64\% & 0.947 & 0.263 & 0.412 \\
  & & $S_{\text{sem}}$   & \textbf{0.042} & 65\% & \textbf{0.969} & 0.269 & 0.421 \\
  & & $S_{\text{prob}}$  & 0.065 & \textbf{23\%} & 0.951 & \textbf{0.295} & \textbf{0.450} \\ \cmidrule(lr){2-8}
& \multirow{4}{*}{Llama-3.2-Vision}
  & CONFLVLM           & 0.105 & 51\% & 0.936 & 0.157 & 0.269 \\
   & & $T_{\text{prob}}$  & 0.045 & 64\% & \textbf{0.973} & 0.154 & 0.266 \\
  & & $S_{\text{sem}}$   & 0.067 & 73\% & 0.967 & 0.151 & 0.262 \\
  & & $S_{\text{prob}}$  & \textbf{0.037} & \textbf{13\%} & 0.940 & \textbf{0.180} & \textbf{0.302} \\
  \cmidrule(lr){2-8}

& \multirow{4}{*}{Qwen2.5-VL-7B}
  & CONFLVLM           & \textbf{0.039} & 42\% & 0.959  & 0.142 & 0.247 \\
   & & $T_{\text{prob}}$   & 0.053  & 53\% & \textbf{0.971} & 0.139 & 0.243 \\
  & & $S_{\text{sem}}$    & 0.075 & 61\% & 0.959 & 0.135 & 0.237 \\
  & & $S_{\text{prob}}$  & 0.045 & \textbf{0\%} & 0.852 & \textbf{0.194} & \textbf{0.316} \\
  \cmidrule(lr){2-8}

& \multirow{4}{*}{Qwen3-VL-8B}
  & CONFLVLM           & 0.063 & 30\% & 0.943 & 0.081 & 0.150 \\
   & & $T_{\text{prob}}$   & \textbf{0.035} & 12\% & \textbf{0.974} & 0.086 & 0.158  \\
  & & $S_{\text{sem}}$    & 0.070 & 43\% & 0.961 & 0.079 & 0.146 \\
  & & $S_{\text{prob}}$  & 0.036 & \textbf{0\%} & 0.794 & \textbf{0.116} & \textbf{0.202} \\

\midrule
\multirow{8}{*}{\makecell{Fine-Grained \\Captioning}} 
& \multirow{4}{*}{LLaVA-1.5}
  & CONFLVLM           & 0.015 & \textbf{61}\% & 0.990 & \textbf{0.412} & \textbf{0.582} \\
  & & $T_{\text{prob}}$  & 0.022 & 83\% & 0.992 & 0.391 & 0.561 \\
  & & $S_{\text{sem}}$   & 0.052 & 91\% & 0.984 & 0.383 & 0.551 \\
  & & $S_{\text{prob}}$  & \textbf{0.003} & 78\% & \textbf{0.997} & 0.398 & 0.569 \\  \cmidrule(lr){2-8}
& \multirow{4}{*}{Phi-3.5-Vision}
  & CONFLVLM           & 0.047 & 63\% & 0.959  & 0.296  & 0.453 \\
  & & $T_{\text{prob}}$  & \textbf{0.030} & 77\% & \textbf{0.991} & 0.296 & 0.456 \\
  & & $S_{\text{sem}}$   & 0.039 & 88\% & 0.989  & 0.291  & 0.450 \\
  & & $S_{\text{prob}}$  & 0.059 & \textbf{8\%} & 0.961 & \textbf{0.346} & \textbf{0.508} \\
  
\midrule
\multirow{8}{*}{\makecell{Document \\Understanding}} 
& \multirow{4}{*}{LLaVA-1.5}
  & CONFLVLM           & 0.073 & 77\% & 0.976 & 0.437 & 0.604 \\
  & & $T_{\text{prob}}$  & 0.070 & 92\% & 0.985 & 0.431 & 0.599 \\
  & & $S_{\text{sem}}$   & \textbf{0.020} & 94\% & \textbf{0.996} & 0.434 & 0.605 \\
  & & $S_{\text{prob}}$  & 0.075 & \textbf{40\%} & 0.978 & \textbf{0.462} & \textbf{0.627} \\  \cmidrule(lr){2-8}
& \multirow{4}{*}{Phi-3.5-Vision}
  & CONFLVLM           & 0.095 & 68\% & 0.970 & 0.239 & 0.384 \\
  & & $T_{\text{prob}}$  & 0.040 & 88\% & 0.987 & 0.239 & 0.385 \\
  & & $S_{\text{sem}}$   & 0.035 & 87\% & 0.990 & 0.243 & 0.390 \\
  & & $S_{\text{prob}}$  & \textbf{0.020} & \textbf{53\%} & \textbf{0.995} & \textbf{0.252} & \textbf{0.402} \\

\bottomrule
\end{tabular}
\vspace{-0.25cm}
\caption{
Conformal risk control on the test sets. The per-$\lambda$ test uses $\alpha = 0.10$, yielding a concurrently guaranteed level $\alpha' = 0.170$ (Section~\ref{sec:crc}). Response-level metrics (empirical risk, abstention) reflect the formal CRC guarantee; claim-level metrics (TPR, precision, F1) are diagnostic. Abstention is a response-level metric: the fraction of responses for which \emph{all} generated claims are filtered by the calibrated threshold (and which incur zero loss under the selective-prediction convention, Section~\ref{sec:crc}), not the fraction of individual claims filtered. CONFLVLM uses CLIP-ViT-Large as its conformity score~\citep{li2025towards}; $T_{\text{prob}}$ is a token-probability baseline; $S_{\text{sem}}$ and $S_{\text{prob}}$ are our proposed conformity signals. All methods satisfy the bound; $S_{\text{prob}}$ achieves the lowest abstention and highest F1 in most settings, while $S_{\text{sem}}$ tends to be more conservative (higher abstention) through stricter filtering.
}
\label{tab:mscoco_main_results}
\vspace{-0.5cm}
\end{table*}

%%%%%%%%%%%%%%%%%%%%%%%%%%%%%%%%%%%%%%%%%%%%%%%%%%%%%%%%%%%%%%%%%%

\begin{figure*}[t]
    \centering

    \begin{subfigure}[t]{0.245\textwidth}
        \centering
        \includegraphics[width=\linewidth]{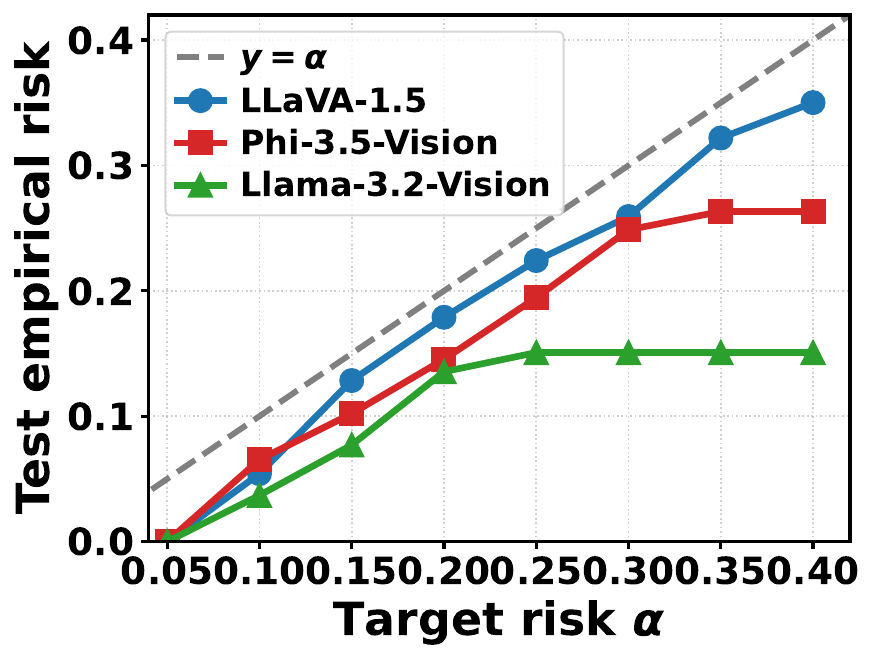}
        \vspace{-0.75cm}
        \caption{}
        \label{fig:mscoco_validity}
    \end{subfigure}
    \hfill
    \begin{subfigure}[t]{0.245\textwidth}
        \centering
        \includegraphics[width=\linewidth]{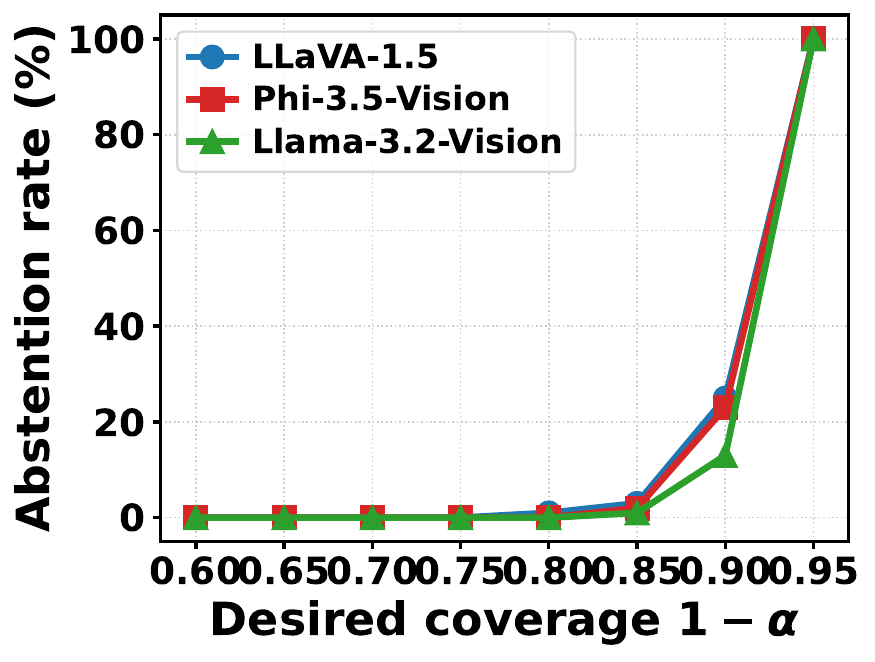}
        \vspace{-0.75cm}
        \caption{}
        \label{fig:mscoco_coverage}
    \end{subfigure}
    \hfill
    \begin{subfigure}[t]{0.245\textwidth}
        \centering
        \includegraphics[width=\linewidth]{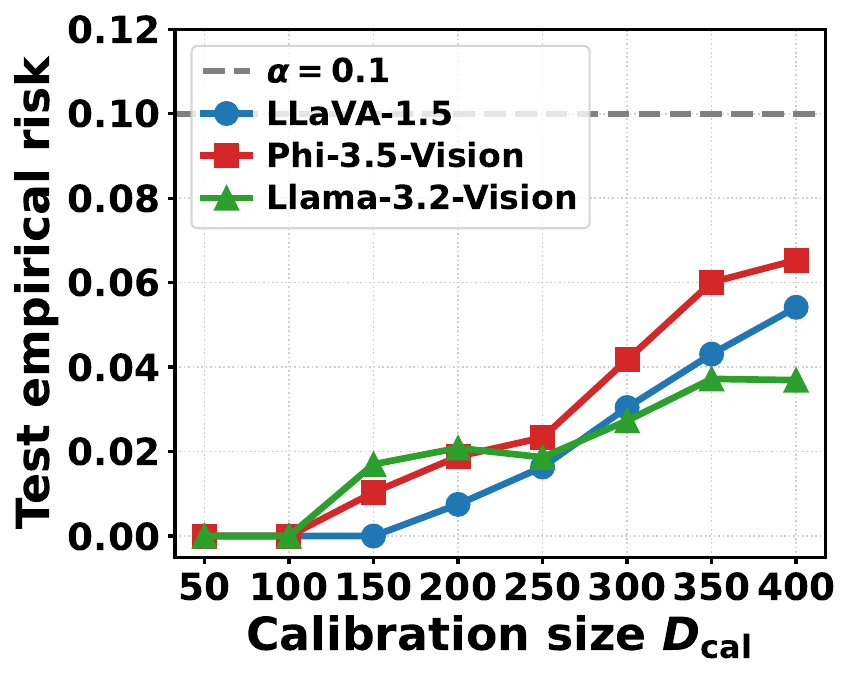}
        \vspace{-0.75cm}
        \caption{}
        \label{fig:mscoco_calsize_risk}
    \end{subfigure}
    \hfill
    \begin{subfigure}[t]{0.245\textwidth}
        \centering
        \includegraphics[width=\linewidth]{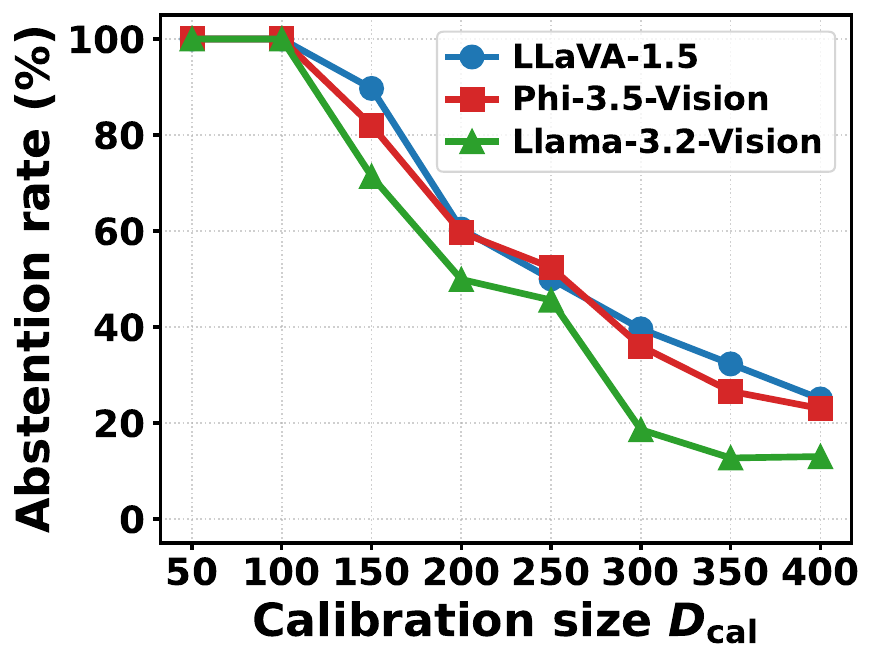}
        \vspace{-0.75cm}
        \caption{}
        \label{fig:mscoco_calsize_abst}
    \end{subfigure}
    \vspace{-0.5cm}
    \caption{
    MSCOCO evaluation across three LVLMs. \textbf{(a)} CRC validity under varying user target $\alpha \in [0.05, 0.40]$: empirical risk remains below the target for all models, confirming the conformal guarantee. \textbf{(b)} Abstention rate as a function of desired coverage $1 - \alpha$, characterizing the utility cost of stricter risk control. \textbf{(c)} Test empirical risk and \textbf{(d)} abstention rate under varying calibration size $D_{\text{cal}}$, at the operating point used throughout (per-$\lambda$ test at $\alpha = 0.10$, concurrently guaranteed level $\alpha' = 0.170$; shown as $\alpha = 0.1$ in the panel legends).
    }
    \label{fig:mscoco_all}
    \vspace{-0.5cm}
\end{figure*}

%%%%%%%%%%%%%%%%%%%%%%%%%%%%%%%%%%%%%%%%%%%%%%%%%%%%%%%%%%%%%%%%%%%

%%%%%--------------------------------------------------

\subsection{Introspective Signal Quality}
Table~\ref{tab:signal_quality_mscoco} evaluates the ability of different conformity signals to distinguish factual from non-factual claims on the calibration set. We compare the external CLIP-based verifier used by CONFLVLM~\citep{li2025towards}, average token probability ($T_{\text{prob}}$), and our proposed introspective signals: layer-wise semantic stability ($S_{\text{sem}}$) and verification probability ($S_{\text{prob}}$). For each signal, we report the mean score on factual and non-factual claims, their difference, AUROC, and Welch's $t$-test $p$-value.

\noindent\textbf{$S_{\text{prob}}$ substantially outperforms external and confidence-based signals.} Across all tasks and LVLM architectures, $S_{\text{prob}}$ consistently achieves the strongest separation between factual and non-factual claims. On MSCOCO, $S_{\text{prob}}$ improves the factual/non-factual score gap from $+0.0196$ (CLIP) and $+0.0523$ ($T_{\text{prob}}$) to $+0.2014$ on LLaVA-1.5, while achieving the highest AUROC of 0.819. Similar trends hold for Phi-3.5-Vision, where $S_{\text{prob}}$ attains a separation of $+0.2397$ and AUROC of 0.763, substantially outperforming the external CLIP verifier used by CONFLVLM. In contrast, generation-time confidence signals ($T_{\text{prob}}$) exhibit substantially weaker discrimination across most settings, suggesting that decoding confidence alone is insufficient for reliable factuality estimation. The baseline signals can in fact be \emph{anti-correlated} with factuality: $T_{\text{prob}}$ on document understanding (LLaVA-1.5) yields a negative gap ($-0.0182$, AUROC $0.493$), and $S_{\text{sem}}$ on Llama-3.2-Vision MSCOCO reverses similarly ($-0.0003$, AUROC $0.488$), assigning higher scores to non-factual claims. These reversals show that neither generation-time confidence nor hidden-state stability is universally reliable across architectures and tasks, motivating the more direct $S_{\text{prob}}$ signal.

\noindent\textbf{$S_{\text{prob}}$ generalizes consistently across tasks and architectures.} The same trend holds beyond scene understanding. On fine-grained captioning, $S_{\text{prob}}$ achieves the highest AUROC across both models (0.765 and 0.770), with large factual/non-factual separations of $+0.1849$ and $+0.2468$. On document understanding, despite the increased difficulty of structured financial documents, $S_{\text{prob}}$ continues to provide the strongest discrimination, reaching AUROC values up to 0.728. In comparison, $S_{\text{sem}}$ alone yields only modest separability, with factual/non-factual score differences often below $+0.002$, though it remains statistically significant in most settings, confirming that hidden-state trajectories carry a weak but consistent factuality signal. Overall, $S_{\text{prob}}$ provides substantially stronger factuality cues than external verification or token confidence across all evaluated settings.

%%%%%-----------------------------------------------

\subsection{Conformal Risk Control Results}

Table~\ref{tab:mscoco_main_results} reports conformal risk control performance on the held-out test sets, using a per-$\lambda$ test at $\alpha = 0.10$ with a concurrently guaranteed level $\alpha' = 0.170$ (Section~\ref{sec:crc}). Following CONFLVLM~\citep{li2025towards}, we evaluate response-level empirical risk and abstention, which correspond directly to the formal CRC guarantee, while claim-level filtering efficiency (TPR), precision, and F1 are reported as diagnostic metrics. We compare IntroConformal against CONFLVLM using its CLIP external verifier and the token-probability baseline $T_{\text{prob}}$.

\noindent\textbf{$S_{\text{prob}}$ satisfies the CRC guarantee with substantially lower abstention and stronger claim-level discrimination.} Across all tasks and LVLM architectures, the proposed conformity signals satisfy the conformal risk requirement, with empirical test risk consistently below the guaranteed level. $S_{\text{sem}}$ is conservative, often yielding low empirical risk at the cost of high abstention, whereas $S_{\text{prob}}$ delivers substantially lower abstention and stronger claim-level performance. On MSCOCO with LLaVA-1.5, $S_{\text{prob}}$ reduces abstention from 57\% (CONFLVLM) and 64\% ($S_{\text{sem}}$) to 25\% while improving F1 from 0.504 to 0.581, and it improves F1 from 0.269 to 0.302 on Llama-3.2-Vision. $S_{\text{prob}}$ achieves the highest F1 in three of four fine-grained captioning and document understanding settings, indicating that stronger signal-level discrimination translates into more efficient conformal filtering while retaining substantially more responses.

Figure~\ref{fig:mscoco_all} further analyzes CRC behavior on MSCOCO across varying target risks and calibration sizes. Figure~\ref{fig:mscoco_validity} shows that empirical risk remains below the desired target across all models and values of $\alpha$, confirming valid finite-sample conformal control. Figure~\ref{fig:mscoco_coverage} illustrates the expected abstention--coverage trade-off, where stricter risk control induces higher abstention. Figures~\ref{fig:mscoco_calsize_risk} and~\ref{fig:mscoco_calsize_abst} sweep the calibration set size from 50 to 400 examples: calibration becomes increasingly efficient as it grows, with empirical risk approaching the target from below and abstention decreasing substantially between 100 and 200 samples before stabilizing. Extended CRC analyses for fine-grained captioning and document understanding are in Appendix~\ref{sec:crc_behavior}.

%%%%--------------------------------------------

\begin{table}[!ht]
    \centering
    \small
    \begin{tabular}{lcc}
        \toprule
        \textbf{Method} & \textbf{\makecell{Claim Filtering\\Efficiency (TPR) $\uparrow$}} & \textbf{\makecell{Response\\Accuracy $\uparrow$}} \\
        \midrule
        Woodpecker &59.1\%  &41\% \\
        CoVe  &37.0\%  &23\% \\
        VCD ($\beta$ = 0.1)  &35.5\%  &20\% \\
        ICD ($\beta$ = 0.1, P) &41.1\% &26\% \\
        \cmidrule(lr){1-3}
        CONFLVLM  & 95.3\% & 90\%\\
        IntroConformal & \textbf{97.4\%} & \textbf{91}\% \\
        \bottomrule
    \end{tabular}
    \vspace{-0.25cm}
    \caption{Comparison with decoding- and verification-based methods on claim filtering efficiency and response accuracy. IntroConformal ($S_{\text{prob}}$) outperforms all baselines, including CONFLVLM, while requiring no external models or decoding-time perturbations.}
    \label{tab:comparison}
    \vspace{-0.5cm}
\end{table}

%%%%---------------------------------------

\subsection{Comparison with Decoding- and Verification-Based Methods}
\label{sec:comparison}
We further compare IntroConformal against representative hallucination mitigation approaches, including Woodpecker~\citep{woodpecker2024}, Chain-of-Verification (CoVe)~\citep{cove2024}, Visual Contrastive Decoding (VCD)~\citep{vcd}, and Instruction Contrastive Decoding (ICD)~\citep{icd2024}. Baseline results are taken directly from CONFLVLM, where all methods were evaluated on the LLaVA-1.5 general scene understanding benchmark using the same 100-image subset and original implementation settings. Response accuracy measures the fraction of responses in which all retained claims are factual.

\noindent\textbf{IntroConformal outperforms all decoding- and verification-based baselines.} As shown in Table~\ref{tab:comparison}, IntroConformal achieves the strongest overall performance, improving claim filtering efficiency from 95.3\% to 97.4\% over CONFLVLM while also achieving slightly higher response accuracy (91\% vs.\ 90\%). It further outperforms Woodpecker, CoVe, VCD, and ICD, all of which exhibit considerably lower filtering efficiency and response accuracy. These results suggest that signals derived directly from the model provide a more reliable basis for factuality control than external verification heuristics or decoding-time perturbation strategies.

%%%%--------------------------------------

\subsection{Robustness to Annotation Noise}
\label{sec:label_noise}
Because the CRC guarantee is defined relative to the calibration labels, we assess how label noise affects calibration. On LLaVA-1.5 MSCOCO, we inject symmetric noise into the calibration labels at $5\%$, $10\%$, and $15\%$ by randomly flipping that fraction of claim labels, recalibrate the threshold on the corrupted labels, and evaluate empirical risk on the held-out test set against the true labels, averaging over $20$ noise draws (Table~\ref{tab:label_noise}).

\begin{table}[!ht]
\centering
\small
\begin{tabular}{cccc}
\toprule
\textbf{Noise} & $\hat{\lambda}$ & \textbf{Test Risk} & \textbf{Abstention} \\
\midrule
$0\%$  & $0.940$ & $0.054$ & $25.0\%$ \\
$5\%$  & $0.956$ & $0.012$ & $53.5\%$ \\
$10\%$ & $0.961$ & $0.001$ & $67.8\%$ \\
$15\%$ & $0.964$ & \textless0.001 & $77.1\%$ \\
\bottomrule
\end{tabular}
\vspace{-0.25cm}
\caption{Robustness to calibration label noise on LLaVA-1.5 MSCOCO. Symmetric noise is injected into the calibration labels, and test risk is measured against the true labels ($20$ draws averaged). The $0\%$ row reproduces the LLaVA-1.5 operating point in Table~\ref{tab:mscoco_main_results} (per-$\lambda$ test at $\alpha = 0.10$, concurrent guarantee $\alpha' = 0.170$). Test risk stays below the target at every noise level.}
\label{tab:label_noise}
\vspace{-1em}
\end{table}

Across all noise levels, the empirical test risk stays below the target $\alpha = 0.10$ and in fact decreases as noise increases, from $0.054$ at $0\%$ noise to below $0.001$ at $15\%$. The mechanism is structural: random flips inflate the apparent risk on the calibration set, so the LTT procedure selects a larger threshold and filters more conservatively, raising abstention (from $25\%$ to $77\%$) rather than violating the bound. The guarantee therefore degrades gracefully under symmetric annotation error, trading utility for continued validity. We note this analysis addresses symmetric noise; systematic annotation bias, which need not inflate apparent risk, could in principle select a permissive threshold, which we flag in the Limitations.

%%%%%%%%%%%%%%%%%%%%%%%%%%%%%%%%%%%%%%%%%%%%%%%%%%%%%%%

\section{Conclusion}
\label{sec:conclusion}

We introduced IntroConformal, a training-free framework for conformal factuality control in LVLMs using introspective signals derived entirely from the model itself. By leveraging layer-wise semantic stability and verification probability, IntroConformal provides finite-sample, distribution-free guarantees on response-level non-factual risk without relying on external verifiers or auxiliary models. Across diverse vision--language generation tasks, $S_{\text{prob}}$ consistently achieves stronger factual/non-factual discrimination than CLIP-based verification and generation-time confidence signals, leading to lower abstention while maintaining valid conformal guarantees. These results indicate that model-internal signals provide a reliable predictive indicator of non-factual generation, and that combining model-derived conformity scores with CRC offers a principled foundation for trustworthy LVLM deployment in safety-critical applications.

%%%%%%%%%%%%%%%%%%%%%%%%%%%%%%%%%%%%%%%%%%%%%%%%%%%%%%%

\section*{Limitations}
$S_{\text{sem}}$ requires white-box access to hidden states, limiting it to architectures that expose internal activations, whereas $S_{\text{prob}}$ needs only output logits at a single position and thus applies to any open-weight model or logit-exposing API, but not to APIs that withhold logits. Both signals require an additional forward pass per claim, comparable in cost to the CLIP scoring used by CONFLVLM. The guarantee is defined relative to the calibration labels rather than to human ground truth: while it is robust to symmetric label noise (Section~\ref{sec:label_noise}), systematic annotation bias could select a permissive threshold, and our human validation used only a single annotator. The reported $\alpha' = 0.170$ is the FWER-corrected bound for a user target of $\alpha = 0.10$, a benchmark demonstration point rather than a deployment recommendation, and the $\alpha$-to-$\alpha'$ gap narrows with calibration size. The guarantee assumes a fixed model under exchangeability, so fine-tuning, RLHF updates, or checkpoint changes (as on versioned APIs) require recalibration; relatedly, since $S_{\text{prob}}$ reads the model's own verification logits, adversarially crafted inputs could bias the Yes/No logits and void the bound, motivating future work on robustifying introspective scores. Finally, as guarantees are probabilistic (holding with probability at least $1-\delta$), safety-critical deployment should retain human oversight.

%%%%%%%%%%%%%%%%%%%%%%%%%%%%%%%%%%%%%%%%%%%%%%%%

\section*{Acknowledgments}
We acknowledge Advanced Research Computing (ARC) at Virginia Tech for providing the computational resources and technical support that contributed to the results reported in this paper. We thank the authors of CONFLVLM for sharing their resources. We also thank the reviewers for their constructive feedback, which helped improve this paper.

\bibliography{references}

%%%%%%%%%%%%%%%%%%%%%%%%%%%%%%%%%%%%%%%%%%%%%%%%%%%%%%%%%%%%%%%%%%%%%%%%%%%%%%%
% APPENDIX
%%%%%%%%%%%%%%%%%%%%%%%%%%%%%%%%%%%%%%%%%%%%%%%%%%%%%%%%%%%%%%%%%%%%%%%%%%%%%%%
\appendix

\section{Appendix}
\label{sec:appendix}

This section discusses the following topics in detail: 

\begin{itemize}[noitemsep]  
    \item Algorithm for IntroConformal (Appendix \ref{sec:algo})
    % \item Probabilistic Bounds on Risk (Appendix \ref{sec:proofs})
    \item Qualitative Examples of IntroConformal (Appendix \ref{sec:example})
    \item CRC Behavior Across Tasks and Calibration Sizes (Appendix \ref{sec:crc_behavior})
    \item Effect of Layer Selection on Semantic Stability (Appendix \ref{sec:effect_layer})
    \item Claim Decomposition and Annotation Prompts (Appendix \ref{sec:prompts})
\end{itemize}

%%%%%%%%%%%%%%%%%%%%%%%%%%%%%%%%%%%%%%%%%%%%%%%%%%%%%%%%%%%%%%%%%%%%

\begin{algorithm}[!ht]
\caption{IntroConformal: Factuality control via introspective signals}
\label{alg:introconformal}
\begin{algorithmic}[1]
\STATE \textbf{Input:} Calibration set $\mathcal{D}_{\text{cal}}=\{(I_i,X_i,\mathcal{C}_i,\mathcal{L}_i)\}_{i=1}^{n}$; risk level $\alpha$; failure probability $\delta$; conformity score $S(\cdot) \in \{S_{\text{sem}}, S_{\text{prob}}\}$.
\STATE \textbf{Output:} Calibrated threshold $\hat{\lambda}$.
\vspace{2pt}
\STATE \textbf{Step 1: Extract introspective signals (single forward pass per claim).}
\FOR{$i=1$ \TO $n$}
    \FOR{each claim $c \in \mathcal{C}_i$}
        \STATE Run LVLM inference with hidden-state outputs on verification prompt.
        \STATE Compute $S_{\text{sem}}(c)$ via cosine similarity of mid- and late-layer hidden states.
        \STATE Compute $S_{\text{prob}}(c)$ via $P(\texttt{Yes}) / (P(\texttt{Yes}) + P(\texttt{No}))$ at answer position.
    \ENDFOR
\ENDFOR
\vspace{2pt}
\STATE \textbf{Step 2: Define candidate thresholds and $\hat{\alpha}$.}
\STATE $\Lambda \gets \{S(c): i\in[n],\, c\in\mathcal{C}_i\}$.
\STATE Optionally augment $\Lambda$ with a value below $\min(\Lambda)$ to represent retaining all claims.
\STATE $\hat{\alpha} \gets \alpha - \left[\sqrt{\frac{\log(m/\delta)}{2n}} - \sqrt{\frac{\log(1/\delta)}{2n}}\right] $
\vspace{2pt}
\STATE \textbf{Step 3: Compute per-response risk for each threshold.}
\FOR{$i=1$ \TO $n$}
    \FOR{each $\lambda \in \Lambda$}
        \STATE $\hat{\mathcal{C}}_{\lambda}(I_i,X_i) \gets \{c\in\mathcal{C}_i : S(c)\ge \lambda\}$.
        \STATE $m_i(\lambda) \gets \left|\hat{\mathcal{C}}_{\lambda}(I_i,X_i)\right|$.
        \IF{$m_i(\lambda) = 0$}
            \STATE $r_i(\lambda) \gets 0$ \hfill \COMMENT{Abstention}
        \ELSE
            \STATE $r_i(\lambda) \gets \frac{1}{m_i(\lambda)} \sum_{c\in\hat{\mathcal{C}}_{\lambda}(I_i,X_i)} \mathcal{L}(c,I_i)$
        \ENDIF
    \ENDFOR
\ENDFOR
\vspace{2pt}
\STATE \textbf{Step 4: Compute aggregate risk and UCB for each threshold.}
\FOR{each $\lambda \in \Lambda$}
    \STATE $\hat{R}(\lambda) \gets \frac{1}{n}\sum_{i=1}^{n} r_i(\lambda)$.
    \STATE $\mathrm{UCB}_{\delta}(\lambda) \gets \hat{R}(\lambda) + \sqrt{\frac{\log(1/\delta)}{2n}}$
\ENDFOR
\vspace{2pt}
\STATE \textbf{Step 5: Select least conservative concurrently valid threshold.}
\STATE $\hat{\lambda} \gets \min\{\lambda\in\Lambda : \mathrm{UCB}_{\delta}(\lambda) \le \hat{\alpha}\}$.
\vspace{2pt}
\STATE \textbf{Return:} $\hat{\lambda}$.
\end{algorithmic}
\end{algorithm}

%%%%%----------------------------------------------
\subsection{Algorithm for IntroConformal}
\label{sec:algo}

Algorithm~\ref{alg:introconformal} summarizes the full IntroConformal pipeline. For each calibration claim, we extract $S_{\text{sem}}$ from hidden-state representations and $S_{\text{prob}}$ from the model's binary verification judgment in a single forward pass. The calibration phase applies the Learn--Then--Test procedure with Hoeffding's inequality to select the least conservative threshold satisfying the target risk $\alpha$. Formal definitions and theoretical guarantees appear in the main text.

%%%%%%%%%%%%%%%%%%%%%%%%%%%%%%%%%%%%%%%%%%%%%%%%%%%%%%%%%%%%%%%

\subsection{Qualitative Examples of IntroConformal}
\label{sec:example}

\begin{figure*}[!ht]
    \centering
    \begin{subfigure}{\linewidth}
        \includegraphics[width=\linewidth]{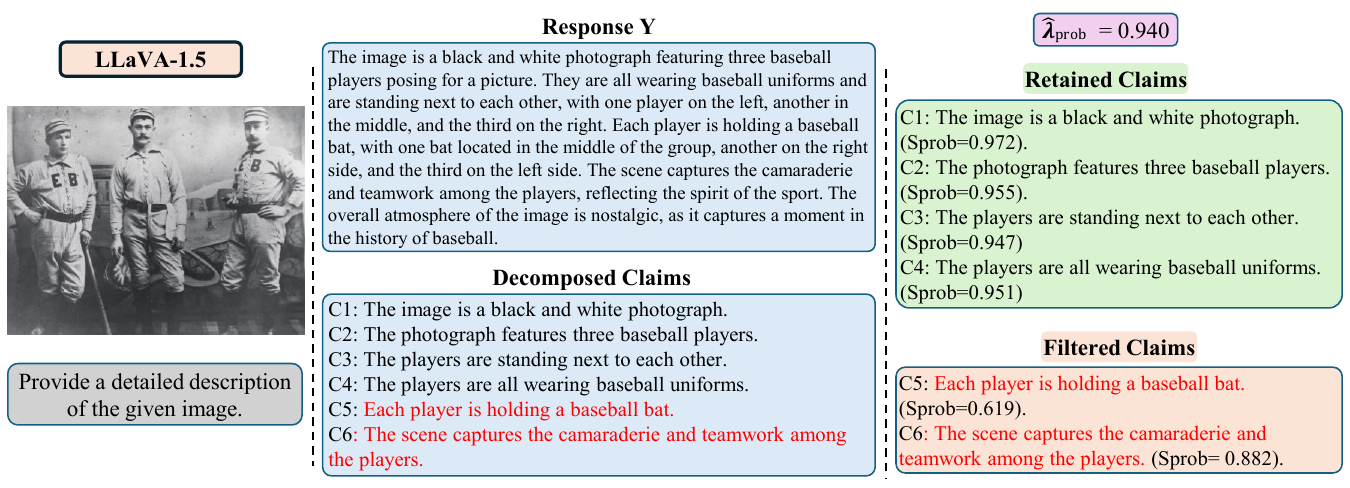}
        \caption{A general scene understanding image whose LVLM response is decomposed into 6 atomic claims. IntroConformal retains all 4 factual claims (green) and correctly filters both non-factual claims (red) using $S_{\mathrm{prob}}$, with $\hat{\lambda}_{\mathrm{prob}} = 0.940$.}
        \label{fig:example1}
    \end{subfigure}
    \vspace{0.2cm}
    \begin{subfigure}{\linewidth}
        \includegraphics[width=\linewidth]{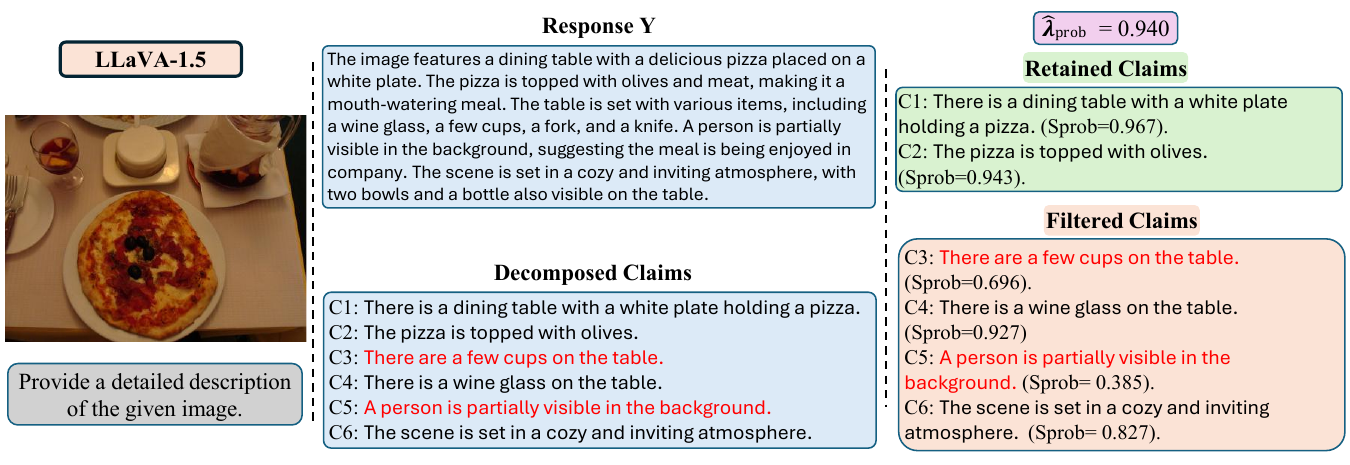}
        \caption{A general scene understanding image whose response is decomposed into 6 atomic claims. IntroConformal correctly filters both non-factual claims (red) while retaining 2 factual claims (green); two additional factual claims are conservatively filtered (black), reflecting the risk--utility trade-off at stringent risk levels, with $\hat{\lambda}_{\mathrm{prob}} = 0.940$.}
        \label{fig:example2}
    \end{subfigure}
    \caption{Qualitative examples of IntroConformal at the operating point used throughout (per-$\lambda$ test at $\alpha = 0.10$, concurrently guaranteed level $\alpha' = 0.170$).}
    \label{fig:examples}
\end{figure*}

\paragraph{IntroConformal correctly filters all non-factual claims.}
Figure~\ref{fig:example1} illustrates a representative case where $S_{\mathrm{prob}}$ achieves perfect claim-level filtering. Given a black-and-white photograph of three baseball players, LLaVA-1.5 generates a response containing two non-factual claims: ``Each player is holding a baseball bat'' ($S_{\mathrm{prob}}=0.619$) and ``The scene captures the camaraderie and teamwork among the players'' ($S_{\mathrm{prob}}=0.882$). Both fall below the calibrated threshold $\hat{\lambda}_{\mathrm{prob}}=0.940$, while all four factual claims score above it ($S_{\mathrm{prob}} \in [0.947, 0.972]$). IntroConformal retains the entire factual set and filters both non-factual claims, yielding a fully grounded response with zero non-factual content among retained claims.

\paragraph{$S_{\mathrm{prob}}$ is sensitive near the decision boundary, reflecting calibrated conservatism.}
Figure~\ref{fig:example2} illustrates the risk--utility trade-off inherent to conformal risk control at stringent target levels. Given a pizza image, LLaVA-1.5 generates 6 claims, two of which are non-factual: ``There are a few cups on the table'' ($S_{\mathrm{prob}}=0.696$) and ``A person is partially visible in the background'' ($S_{\mathrm{prob}}=0.385$); both are correctly filtered. The factual claim ``There is a wine glass on the table'' ($S_{\mathrm{prob}}=0.927$) falls just below $\hat{\lambda}_{\mathrm{prob}}=0.940$, reflecting the sensitivity of the decision boundary where $S_{\mathrm{prob}}$ is close to the threshold, and the borderline claim ``The scene is set in a cozy and inviting atmosphere'' ($S_{\mathrm{prob}}=0.827$) is likewise filtered. This conservative filtering is expected at the operating point (per-$\lambda$ $\alpha=0.10$, guaranteed $\alpha'=0.170$) and is a principled consequence of the guarantee: the threshold is set to bound the non-factual rate among retained claims, which necessarily filters some borderline claims.

%%%%%%%%%%%%%%%%%%%%%%%%%%%%%%%%%%%%%%%%%%%%%%%%%%%%%%%%%%%%%%%%%%

\subsection{CRC Behavior Across Tasks and Calibration Sizes}
\label{sec:crc_behavior}

\begin{figure*}[t]
    \centering

    \begin{subfigure}[t]{0.245\textwidth}
        \centering
        \includegraphics[width=\linewidth]{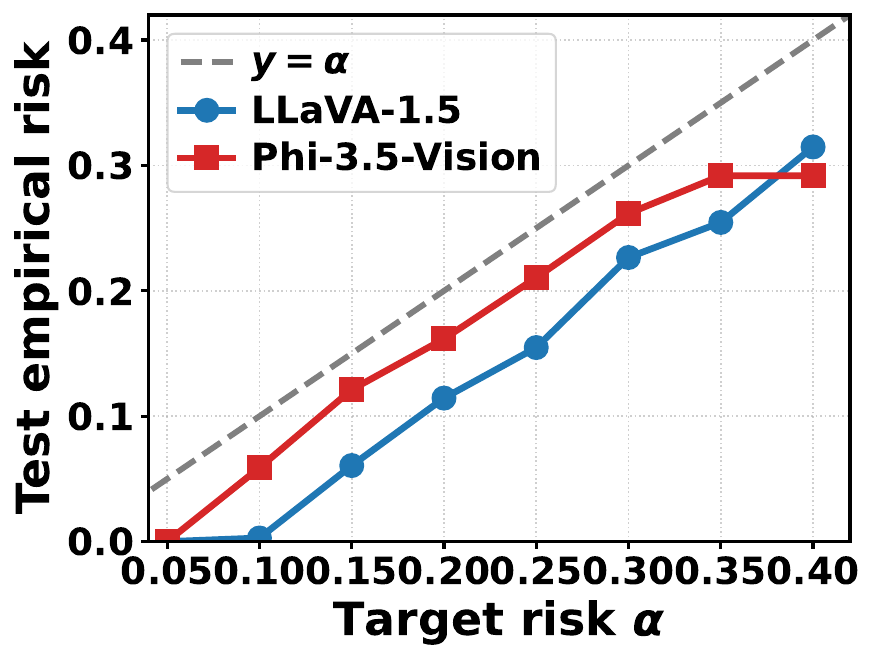}
        \vspace{-0.75cm}
        \caption{}
        \label{fig:fine_grained_validity}
    \end{subfigure}
    \hfill
    \begin{subfigure}[t]{0.245\textwidth}
        \centering
        \includegraphics[width=\linewidth]{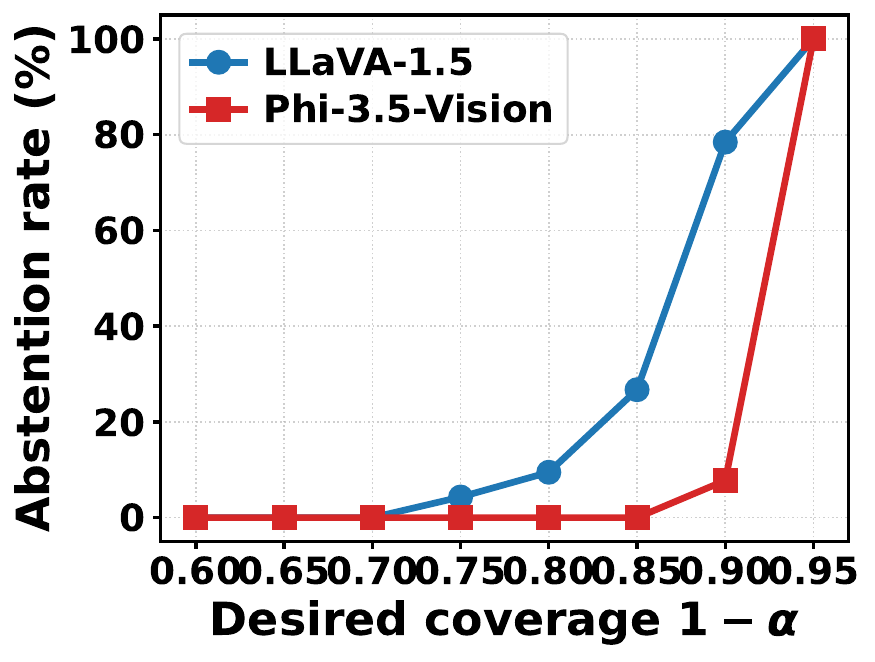}
        \vspace{-0.75cm}
        \caption{}
        \label{fig:fine_grained_coverage}
    \end{subfigure}
    \hfill
    \begin{subfigure}[t]{0.245\textwidth}
        \centering
        \includegraphics[width=\linewidth]{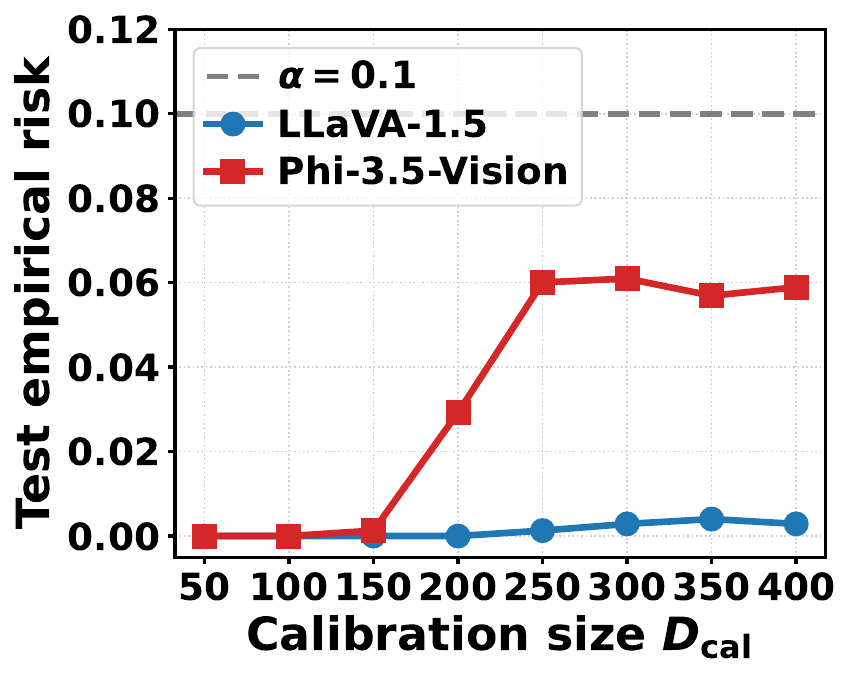}
        \vspace{-0.75cm}
        \caption{}
        \label{fig:fine_grained_calsize_risk}
    \end{subfigure}
    \hfill
    \begin{subfigure}[t]{0.245\textwidth}
        \centering
        \includegraphics[width=\linewidth]{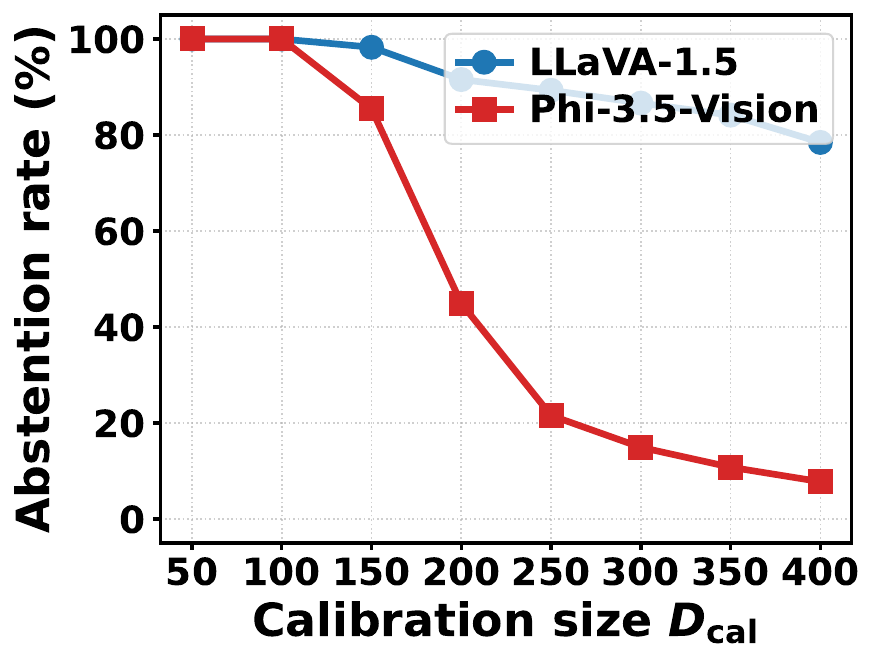}
        \vspace{-0.75cm}
        \caption{}
        \label{fig:fine_grained_calsize_abst}
    \end{subfigure}
    \vspace{-0.5cm}
    \caption{
    Fine-Grained Captioning evaluation across two LVLM architectures. \textbf{(a)} CRC validity under varying user target $\alpha$: empirical risk remains below the target for all models, confirming the conformal guarantee. \textbf{(b)} Abstention rate as a function of desired coverage $1 - \alpha$, characterizing the utility cost of stricter risk control. \textbf{(c)} Test empirical risk and \textbf{(d)} abstention rate under varying calibration size $D_{\text{cal}}$ (swept 50--400), at the operating point used throughout (per-$\lambda$ test at $\alpha = 0.10$, concurrently guaranteed level $\alpha' = 0.170$; shown as $\alpha = 0.1$ in the panel legends).
    }  
    \label{fig:fine_grained_all}
    %\vspace{-0.5cm}
\end{figure*}

\begin{figure*}[t]
    \centering

    \begin{subfigure}[t]{0.245\textwidth}
        \centering
        \includegraphics[width=\linewidth]{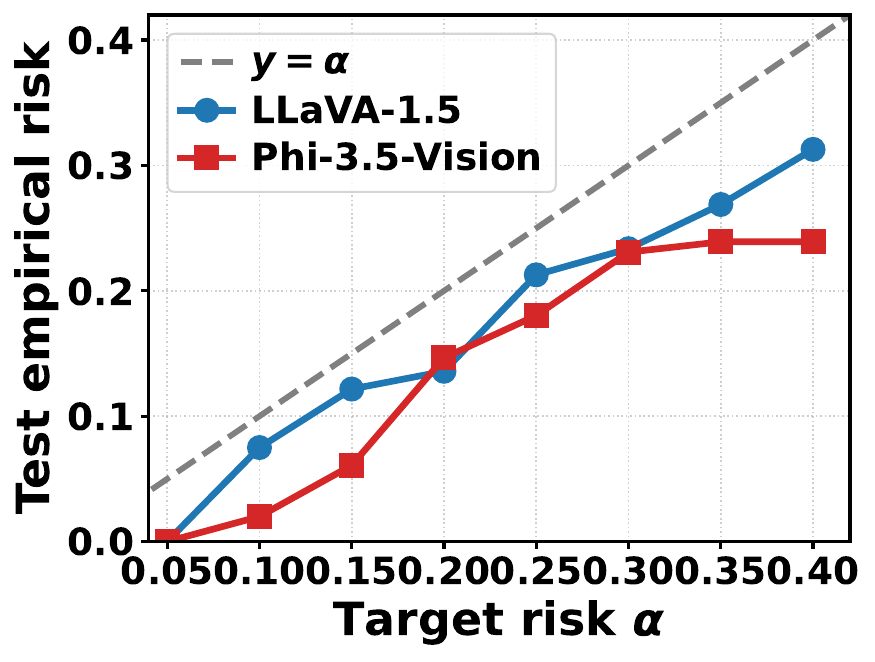}
        \vspace{-0.75cm}
        \caption{}
        \label{fig:doc_validity}
    \end{subfigure}
    \hfill
    \begin{subfigure}[t]{0.245\textwidth}
        \centering
        \includegraphics[width=\linewidth]{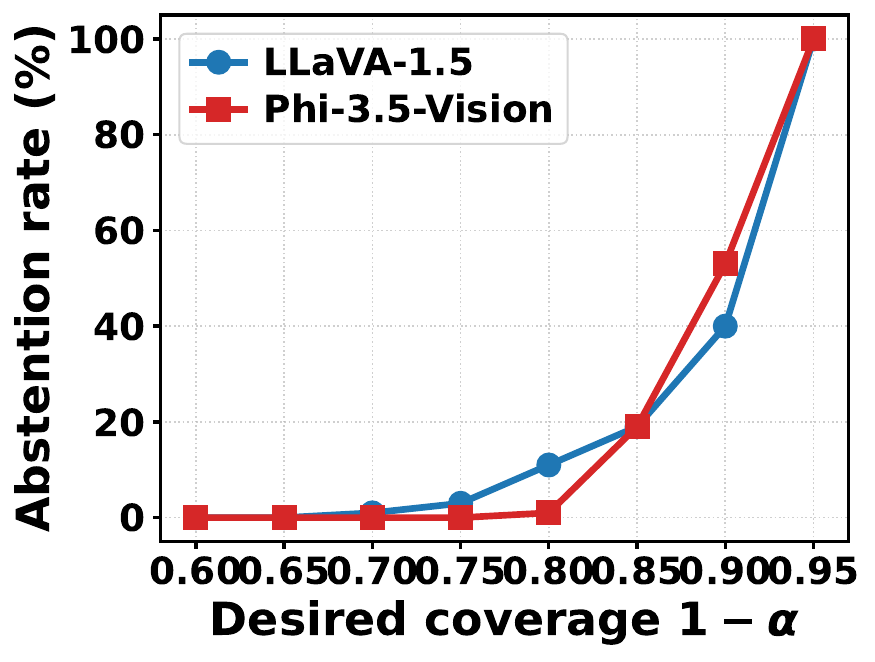}
        \vspace{-0.75cm}
        \caption{}
        \label{fig:doc_coverage}
    \end{subfigure}
    \hfill
    \begin{subfigure}[t]{0.245\textwidth}
        \centering
        \includegraphics[width=\linewidth]{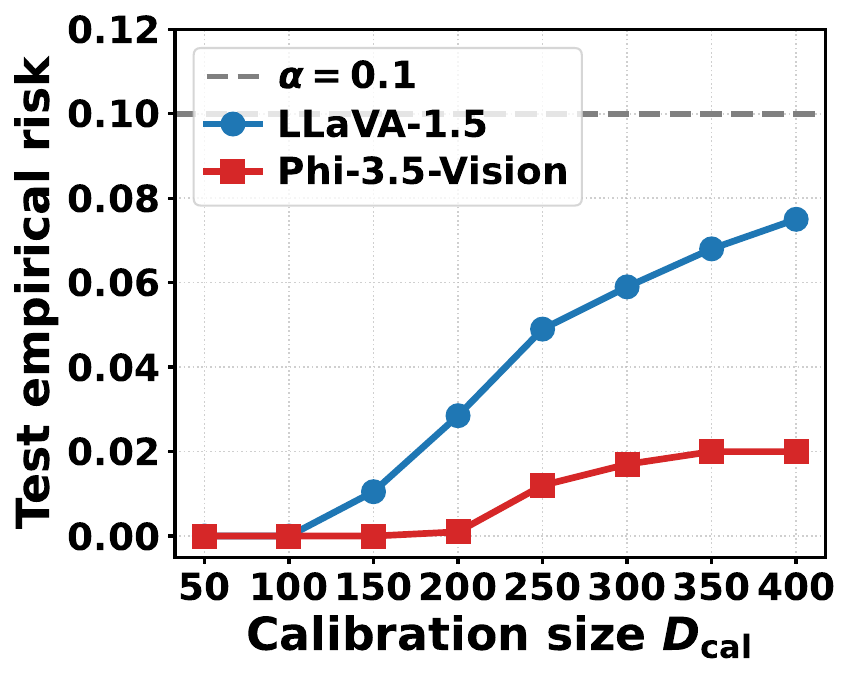}
        \vspace{-0.75cm}
        \caption{}
        \label{fig:doc_calsize_risk}
    \end{subfigure}
    \hfill
    \begin{subfigure}[t]{0.245\textwidth}
        \centering
        \includegraphics[width=\linewidth]{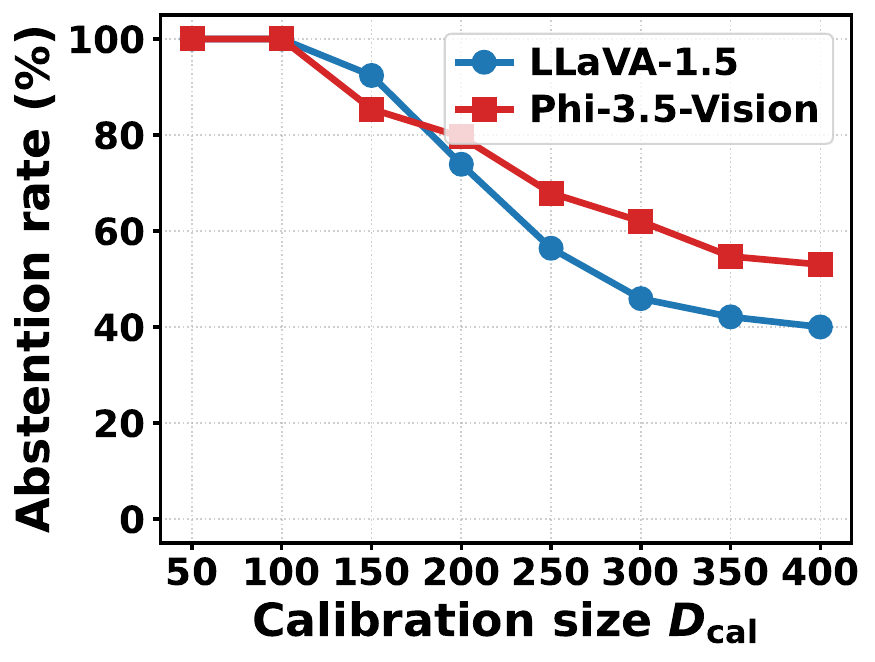}
        \vspace{-0.75cm}
        \caption{}
        \label{fig:doc_calsize_abst}
    \end{subfigure}
    \vspace{-0.5cm}
    \caption{
    Document understanding evaluation across two LVLM architectures. \textbf{(a)} CRC validity under varying user target $\alpha$: empirical risk remains below the target for all models, confirming the conformal guarantee. \textbf{(b)} Abstention rate as a function of desired coverage $1 - \alpha$, characterizing the utility cost of stricter risk control. \textbf{(c)} Test empirical risk and \textbf{(d)} abstention rate under varying calibration size $D_{\text{cal}}$ (swept 50--400), at the operating point used throughout (per-$\lambda$ test at $\alpha = 0.10$, concurrently guaranteed level $\alpha' = 0.170$; shown as $\alpha = 0.1$ in the panel legends).
    }  
    \label{fig:doc_all}
    % \vspace{-0.5cm}
\end{figure*}

Figures~\ref{fig:fine_grained_all} and~\ref{fig:doc_all} extend the CRC analysis from the main text to fine-grained captioning and document understanding, respectively. Across both tasks and all evaluated architectures, the conformal guarantee holds consistently: empirical risk remains below the target $\alpha$ for all values in $[0.05, 0.40]$, confirming valid finite-sample risk control (Figures~\ref{fig:fine_grained_validity} and~\ref{fig:doc_validity}). The abstention--coverage trade-off (Figures~\ref{fig:fine_grained_coverage} and~\ref{fig:doc_coverage}) follows the expected monotonic pattern, where stricter coverage requirements induce higher abstention, with Phi-3.5-Vision exhibiting a sharper abstention increase at high coverage thresholds compared to LLaVA-1.5.

Figures~\ref{fig:fine_grained_calsize_risk}, \ref{fig:fine_grained_calsize_abst}, \ref{fig:doc_calsize_risk}, and~\ref{fig:doc_calsize_abst} analyze the effect of calibration size $|\mathcal{D}_{\text{cal}}|$, swept from 50 to 400 examples, at the operating point used throughout (per-$\lambda$ test at $\alpha = 0.10$, concurrently guaranteed level $\alpha' = 0.170$). On fine-grained captioning, LLaVA-1.5 achieves near-zero empirical risk even at small calibration sizes, reflecting the stronger intrinsic signal quality on this task, while Phi-3.5-Vision requires larger calibration sets before risk stabilizes. On document understanding, both models show a consistent decrease in abstention as $|\mathcal{D}_{\text{cal}}|$ grows, with abstention plateauing beyond $|\mathcal{D}_{\text{cal}}| \approx 200$, consistent with the $\mathcal{O}(1/\sqrt{n})$ shrinkage of the Hoeffding upper confidence bound. These results confirm that the calibration efficiency observed on MSCOCO in the main text generalizes across tasks, and that approximately 200 calibration samples suffice for stable conformal risk control in practice.

%%%%%%%%%%%%%%%%%%%%%%%%%%%%%%%%%%%%%%%%%%%%%%%%%%%%%%%%%%%%%%%%%%

%%%%%%%%%%%%%%%%%%%%%%%%%%%%%%%%%%%%%%%%%%%%%%%%%%%%%%%%%%%%%%%%%%%%%

\subsection{Effect of Layer Selection on Semantic Stability}
\label{sec:effect_layer}
Table~\ref{tab:ssem_layer_ablation} compares two layer selection strategies for computing the semantic stability score. The old configuration compares hidden-state representations from the first quarter to the network midpoint against the final quarter of layers, following earlier mechanistic interpretability work~\citep{azaria2023internal, cheninside}. The new configuration instead compares the 8 layers immediately preceding the final block against the final 4 layers, motivated by recent observations that factual representations tend to stabilize in late-stage hidden-state trajectories, where the model commits to its final output~\citep{wang2025mllm, bu2026sampling}. While the old configuration achieves higher AUROC in several settings (e.g., 0.578 vs.\ 0.556 on general scene understanding with LLaVA-1.5), it exhibits two failure modes. First, on fine-grained captioning with LLaVA-1.5, the directional difference is negative ($-0.0029$), meaning the old configuration assigns higher scores to non-factual than factual claims, reversing the intended ordering. Second, on fine-grained captioning with Phi-3.5-Vision, it fails to reach significance ($p = 2.1 \times 10^{-1}$), indicating no reliable separation. We acknowledge that the old configuration achieves notably higher AUROC on document understanding with Phi-3.5-Vision (0.628 vs.\ 0.530), a gap that warrants attention; however, we prioritize cross-architecture consistency over per-setting AUROC maximization, as a score with reversed or unreliable ordering in some settings cannot serve as a dependable conformity score. The new configuration yields consistent directional separation with statistically significant results across the tasks and architectures in this ablation (LLaVA-1.5 and Phi-3.5-Vision), and we adopt it as our default throughout all experiments.

\begin{table*}[!ht]
\centering
\small
\setlength{\tabcolsep}{4pt}
\begin{tabular}{llcccccc}
\toprule
Task & Model & Config & Mean (F) & Mean (NF) & Diff. & AUROC $\uparrow$ & $p$-value \\
\midrule
\multirow{4}{*}{\shortstack[l]{General Scene\\Understanding}}
 & \multirow{2}{*}{LLaVA-1.5}
   & Old & 0.4952 & 0.4887 & +0.0065 & \textbf{0.578} & $2.7 \times 10^{-13}$ \\ % Done
 & & New & 0.8689 & 0.8674 & +0.0015 & 0.556 & $3.1 \times 10^{-9}$ \\
\cmidrule{2-8}
 & \multirow{2}{*}{Phi-3.5-Vision}
   & Old & 0.2710 & 0.2618 & +0.0092 & \textbf{0.601} & $6.2 \times 10^{-26}$ \\ % Done
 & & New & 0.9041 & 0.9022 & +0.0019 & 0.576 & $2.2 \times 10^{-10}$ \\
\midrule
\multirow{4}{*}{\shortstack[l]{Fine-Grained\\Captioning}}
 & \multirow{2}{*}{LLaVA-1.5}
   & Old & 0.4921 & 0.4950 & -0.0029 & 0.471 & $6.2 \times 10^{-4}$ \\ % Done
 & & New & 0.8703 & 0.8695 & +0.0008 & \textbf{0.536} & $3.7 \times 10^{-4}$ \\
\cmidrule{2-8}
 & \multirow{2}{*}{Phi-3.5-Vision}
   & Old & 0.2772 & 0.2765 & +0.0008 & 0.520 & $2.1 \times 10^{-1}$ \\
 & & New & 0.9033 & 0.9025 & +0.0008 & \textbf{0.530} & $6.4 \times 10^{-3}$ \\
\midrule
\multirow{4}{*}{\shortstack[l]{Document\\Understanding}}
 & \multirow{2}{*}{LLaVA-1.5}
   & Old & 0.4846 & 0.4814 & +0.0032 & 0.555 & $4.7 \times 10^{-10}$ \\ % Done
 & & New & 0.8710 & 0.8691 & +0.0018 & \textbf{0.575} & $5.0 \times 10^{-21}$ \\
\cmidrule{2-8}
 & \multirow{2}{*}{Phi-3.5-Vision}
   & Old & 0.2458 & 0.2364 & +0.0094 & \textbf{0.628} & $3.9 \times 10^{-55}$ \\ % Done
 & & New & 0.8837 & 0.8818 & +0.0019 & 0.530 & $5.7 \times 10^{-6}$ \\
\bottomrule
\end{tabular}
\caption{Effect of $S_{\mathrm{sem}}$ layer selection on signal quality. 
\textbf{Old} takes $\mathcal{M}$ to span the first quarter to the network midpoint and $\mathcal{T}$ the final quarter. \textbf{New} takes $\mathcal{M}$ to be the 8 layers preceding the final block and $\mathcal{T}$ the final 4 layers, following \citet{wang2025mllm} and \citet{bu2026sampling}. Results are reported on the calibration set for LLaVA-1.5 and Phi-3.5-Vision.}
\label{tab:ssem_layer_ablation}
\end{table*}

%%%%%%%%%%%%%%%%%%%%%%%%%%%%%%%%%%%%%%%%%%%%%%%%%%%%%%%%%%%%%%%%%%%

\subsection{Claim Decomposition and Annotation Prompts}
\label{sec:prompts}

We provide the full prompts used for claim decomposition and factuality annotation. The annotation models and reliability study are described in Section~\ref{sec:experi_setup}; here we give the exact prompt text. The claim decomposition prompt and the fine-grained captioning annotation prompt are shown below, followed by the document understanding annotation prompt. The three share an identical claim-level JSON output format; the document prompt differs only in its error taxonomy, which covers field misinterpretation, numerical and quantitative errors, date errors, item errors, and OCR or layout issues.

\begin{figure*}[ht]
\centering
\begin{tcolorbox}[
    title=Claim decomposition prompt,
    fonttitle=\bfseries,
    colback=promptbg,
    colframe=promptborder,
    colbacktitle=promptborder,
    coltitle=white,
    arc=3pt,
    boxrule=0.8pt,
    width=\textwidth
]
\textbf{System} ``Given a model-generated response about an image, decompose it into a list of atomic, verifiable claims.''

\medskip
\textbf{User} ``Statement: \{response\}

Break down the above statement into a list of atomic claims. Each claim must:
\begin{itemize}[leftmargin=1.5em, topsep=2pt, itemsep=1pt]
    \item Express exactly one fact (no compound claims joined by `and'/`or')
    \item Be self-contained, with all referents resolved (no pronouns like `it', `they', `this')
    \item Be directly supported by the original statement (do not add inferences or interpretations)
    \item Ensure all distinct facts from the original statement are represented
    \item Be a short, declarative sentence
    \item Omit interpretive or evaluative claims that cannot be verified from the image alone (e.g., `the scene is dynamic', `a majestic sight', `a serene backdrop')
    \item If a claim contains a hedge (`possibly', `appears to', `seems to', `might be'), restate it as a direct assertion without the hedge
\end{itemize}
Output only a numbered list in the format:\\
1. $\langle$claim$\rangle$\\
2. $\langle$claim$\rangle$\\
$\ldots$

Do not include any explanation or preamble.''
\end{tcolorbox}
\end{figure*}

%%%%----------------------------------------------------

\begin{figure*}[!ht]
\centering
\begin{tcolorbox}[
    title=Factuality annotation prompt for Fine-Grained Captioning,
    fonttitle=\bfseries,
    colback=promptbg,
    colframe=promptborder,
    colbacktitle=promptborder,
    coltitle=white,
    arc=3pt,
    boxrule=0.8pt
]
\textbf{System} ``You are an expert annotator tasked with evaluating statements generated by a vision-language model (VLM). Given an image and a list of claims, verify the factuality of each claim based on how well it aligns with the provided image. Focus only on significant or material correctness, ignoring minor differences or non-essential details.

The errors are categorized as follows:
\begin{enumerate}[leftmargin=1.5em, topsep=2pt, itemsep=1pt]
    \item \textbf{Object Identification}: The claim involves hallucinated or wrongly identified objects, including species, breed, or model misidentification (e.g., wrong bird species, wrong car model, wrong dog breed). Ignore minor distinctions between similar objects unless it fundamentally changes the meaning of the claim.
    \item \textbf{Attribute Accuracy}: The claim involves incorrect visual attributes (e.g., color, size, shape, markings, body parts). Only flag if critical to the understanding of the claim.
    \item \textbf{Spatial Relations}: The claim involves incorrect spatial relationships between objects. Only flag if they significantly change the scene.
    \item \textbf{Interaction/Action Accuracy}: The claim involves incorrect or hallucinated action or interaction.
    \item \textbf{Quantitative Information}: The claim involves incorrect numeric details (e.g., wrong object count).''
\end{enumerate}

\medskip
\textbf{User} ``Given the image and the caption below, determine whether each claim is supported by the image.

Caption: \{caption\}

Claims: \{numbered\_claims\}

For each claim, assign:
\begin{itemize}[leftmargin=1.5em, topsep=2pt, itemsep=1pt]
    \item \textbf{true}: the claim is factually correct and supported by visible evidence in the image
    \item \textbf{false}: the claim is incorrect, hallucinated, or not visually verifiable
\end{itemize}

The output list length must exactly match the number of input claims. Preserve the exact order of the input claims.

Output format (one boolean per claim, in order):\\
\texttt{\{"labels": [true, false, true, ...]\}}''
\end{tcolorbox}
\end{figure*}

\begin{figure*}[t]
\centering
\begin{tcolorbox}[width=\linewidth, colback=gray!5, colframe=black!50,
  title=Factuality annotation prompt for Document Understanding, fonttitle=\bfseries]
\textbf{System} ``You are an expert annotator tasked with evaluating statements generated by a Document AI model. Given a document image and a list of claims, verify the factuality of each claim based on how well it aligns with the provided document.

The errors are categorized as follows:
\begin{enumerate}[leftmargin=*, itemsep=1pt, topsep=2pt]
\item Field Misinterpretation: Incorrectly identifying important fields such as mistaking ``Subtotal'' for ``Total Amount'', or misrecognizing non-existing fields.
\item Numerical and Quantitative Errors: Incorrect amounts, totals, or quantity values, as well as calculation discrepancies (e.g., subtotal, tax, and total relationship).
\item Date Error: Misrecognizing date or misinterpreting date formats.
\item Item Error: Misrecognizing item or item details, or falsely identifying non-existing items.
\item Other Errors: Other errors such as misspelling or misrecognizing characters, layout, and alignment issues.
\end{enumerate}
Return ONLY valid JSON with no explanation and no markdown.''

\vspace{4pt}
\textbf{User} ``Given the document image and the claims below, determine whether each claim is supported by the document.

Claims: \{numbered\_claims\}

For each claim, assign:
\begin{itemize}[leftmargin=*, itemsep=1pt, topsep=2pt]
\item true: the claim is factually correct and supported by visible content in the document
\item false: the claim is incorrect, hallucinated, or not verifiable from the document
\end{itemize}
The output list length must exactly match the number of input claims. Preserve the exact order of the input claims.

Output format: \{``labels'': [true, false, true, ...]\}''
\end{tcolorbox}
\caption{Factuality annotation prompt for the document understanding task (SROIE). The claim decomposition and output format are identical to the other tasks; only the error taxonomy is task-specific.}
\label{fig:doc_prompt}
\end{figure*}

%%%%%%%%%%%%%%%%%%%%%%%%%%%%%%%%%%%%%%%%%%%%%%%%%%%%%%%%%%%%%

\end{document}